\documentclass[journal]{IEEEtran}
\usepackage{amsthm}
\usepackage{comment}
\usepackage{wrapfig}
\usepackage{booktabs}  
\usepackage{eqparbox}  
\usepackage{graphicx}
\usepackage{multirow}
\usepackage{textcomp}
\usepackage{epsfig} 
\usepackage{mathptmx} 
\usepackage{times} 
\usepackage{amsmath} 
\usepackage{amssymb}  
\usepackage{amsfonts}

\usepackage{lipsum}
\usepackage{tabularx}
\usepackage{rotating}
\usepackage{tablefootnote}
\usepackage[hyperfootnotes=true]{hyperref}
\usepackage{float}
\usepackage{picture}
\usepackage[symbol]{footmisc}
\usepackage{afterpage}
\usepackage{textcomp}
\usepackage{siunitx}
\usepackage{textcomp}
\usepackage{colortbl}
\usepackage{expl3}
\usepackage{xcolor}
\usepackage{gensymb}
\usepackage{notoccite}
\usepackage{soul}

\usepackage[LGR,T1]{fontenc} 
\usepackage[utf8]{inputenc}   

\newcommand{\textgreek}[1]{\begingroup\fontencoding{LGR}\selectfont#1\endgroup}
\newtheorem{definition}{Definition}

\definecolor{mygray}{gray}{0.9}
\definecolor{myred}{HTML}{c12322}
\definecolor{myblue}{HTML}{0b6089}
\definecolor{mygreen}{HTML}{c16b04}
\definecolor{mygreen}{HTML}{217844}

\ifCLASSINFOpdf
\else
\fi
\graphicspath{{./02_Figures/}} 

 \newcommand{\etal}{\textit{et al.}}

\begin{document}

%
\title{\color{black}Soft robotic suits: State of the art, core technologies and open challenges}

%

\author{Michele Xiloyannis,~\IEEEmembership{Member,~IEEE,}
        Ryan Alicea,~\IEEEmembership{Student Member,~IEEE,}
        Anna-Maria Georgarakis,~\IEEEmembership{Student Member,~IEEE,}
        Florian L. Haufe,
        Peter Wolf,\\
        Lorenzo Masia$^{*}$~\IEEEmembership{Member,~IEEE,} and
        Robert Riener$^{*}$~\IEEEmembership{Member,~IEEE.}
\thanks{M. Xiloyannis, F. Haufe, A.-M. Georgarakis, P. Wolf and R. Riener are with the Sensory-Motor Systems (SMS) Lab, Institute of Robotics and Intelligent Systems (IRIS), ETH Zurich, Switzerland and the Spinal Cord Injury Center, University Hospital Balgrist, University of Zurich, Zurich, Switzerland. R. Alicea and L. Masia are with the Institut f\"ur Technische Informatik (ZITI), Heidelberg University, Heidelberg, Germany. $^{*}$ The authors contributed equally to this work. This work was supported by the Swiss National Center of Competence in Research (NCCR) Robotics.}%
}

%
%

\markboth{ Soft robotic suits: State of the art, core technologies and open challenges}%
{Xiloyannis \MakeLowercase{\textit{et al.}}: Soft robotic exosuits}
%



\maketitle

\begin{abstract}
Wearable robots are undergoing a disruptive transition, from the rigid machines that populated the science-fiction world in the early eighties to lightweight robotic apparel, hardly distinguishable from our daily clothes. In less than a decade of development, {\color{black} soft robotic suits} have achieved important results in human motor assistance and augmentation. 
In this paper, we start by giving a definition of {\color{black} soft robotic suits} and proposing a taxonomy to classify existing systems. We then critically review the modes of actuation, the physical human-robot interface and the intention-detection strategies of state of the art soft { \color{black} robotic suits}, highlighting the advantages and limitations of different approaches. Finally, we discuss the impact of this new technology on human movements, for both augmenting human function and supporting motor impairments, and identify areas that are in need of further development. 
\end{abstract}


%
\IEEEpeerreviewmaketitle

\section{Introduction}\label{background}
\IEEEPARstart{I}{n} the book ``The Steam Man of the Prairies'', 1868, Edward Ellis describes a gigantic iron humanoid robot, capable of running at superhuman speeds, controlled by its creator, sitting in a carriage right behind it \cite{Edward1869}. It was shortly after the first industrial revolution and the promises of steam power fueled the imagination of science-fiction writers. The technological developments of the following decades were accompanied by equally-impressive visions of robotic devices that could be worn by human beings, to enhance their strength and eventually, save the human race from alien invasions \cite{Scott1979}. 

\begin{figure*}[htbp]
    \centering
    \includegraphics[width = .5\textwidth]{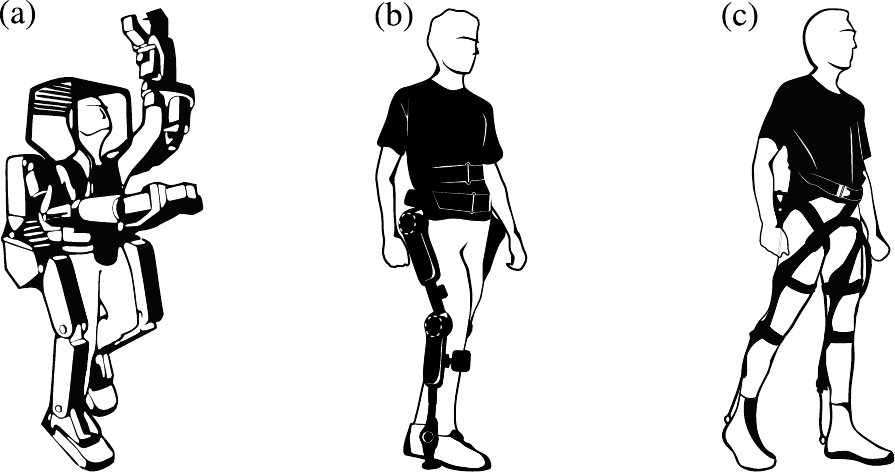}
    \caption{From the Hardiman to soft robotic suits. (a) Concept of the Hardiman, the exoskeleton proposed by Mosher \cite{Mosher1967}, in 1967, with the goal of augmenting human strength. The device relied on a structural metal frame to transfer loads to the ground and weighed \SI{348}{\kilogram}. Image adapted from \cite{Mosher1967}. (b) HAL lower limb exoskeleton (Cyberdyne), 2005. The autonomous exoskeleton weighs \SI{15}{\kilogram}. Image adapted from \cite{Cyberdyne2020}. (c) The Harvard exosuit, first published in 2013 \cite{Asbeck2013}, weighs \SI{5}{\kilogram} in its latest implementation \cite{Kim2019}. Image adapted from \cite{Kusek2014}.}
    \label{Hardi_to_suit}
\end{figure*}

Being inspired by the machines developed to automate the manufacturing industry, the first concepts of wearable robots shared an unsurprising common trait: the frame of the device, roughly reflecting the skeleton of its pilot, was made of rigid metal links connected by mechanical joints. General Electric famously materialized this vision in 1967, when the program led by Ralph Mosher constructed the Hardiman (shown in Figure~\ref{Hardi_to_suit}.a), a \textit{``... coordinated man-machine system [that] will exploit the union of man's superbly integrated sensory system with the tremendous power potential of machinery''} \cite{Mosher1967}. The Hardiman captured the attention and imagination of science-fiction enthusiasts worldwide. Unfortunately, limited by the technology of its time, the robot was never tested with a human pilot. {\color{black} In spite of its failure, this project had the merit of sparking a lasting interest in robots that could enhance or assist human motor performance, through `` ...a mechanical structure that mapped to the human actor's anatomy'' \cite{Pons2008}; it was the birth of exoskeletons.}

Nearly fifty years later, exoskeletons have benefited from the rapid technological advancements in electronics, control, manufacturing, and power storage solutions \cite{Guizzo2005a}. Exoskeletons have successfully augmented human strength during locomotion \cite{Kazerooni2005a}, reduced the metabolic cost of walking \cite{Mooney2014}, restored ambulatory capabilities of paraplegic patients \cite{Kawamoto2002a} (used robot shown in Figure~\ref{Hardi_to_suit}.b), assisted the rehabilitation of stroke patients \cite{Klamroth-Marganska2014}, harvested energy from human movements \cite{Donelan2008}, and helped the study of fundamental principles underlying human motor control \cite{Lam2005}.

Despite their remarkable achievements, exoskeletons are still largely confined to research laboratories or used for expensive clinical treatments only available in elite rehabilitation facilities. The greatest culprit for their limited accessibility possibly lies in their complexity, responsible for increasing cost, weight and size. Part of this complexity stems from the need of a rigid exoskeleton to accurately accompany the intricate motions of the human limbs and joints. Misalignment between the joints of the exoskeleton and those of its user is known to cause uncontrolled interaction forces, which negatively affect the wearer's mobility and metabolics \cite{VanDijk2011} and induces discomfort \cite{Jarrasse2012a}. 

In a recent insight on the evolution of wearable technology, J. L. Pons elegantly pointed out how wearable robots are shifting towards less restrictive and more bio-mimetic architectures, favouring compliant materials that better conform to the anatomical complexity of the human body \cite{Pons2019}.

Pons's vision is justified by the characteristics of the wearable robots developed in the last decade. Capitalizing on the advancement in soft materials, compliant control and non-linear modelling \cite{Laschi2014,Sanchez2020a}, an increasing number of research groups in academia and industry are designing wearable robots made of textiles and elastomers rather than rigid links. This allows the devices to be portable and to assist movements without restricting human biomechanics. {\color{black} This new class of devices has been referred to using different terms, including ``exomuscles'' \cite{Simpson2020}, ``soft exoskeletons''  \cite{Yap2015a} and ``exosuits'' \cite{Asbeck2014b}.}
{\color{black} The term ``suit'' is here used with a different connotation from the one first introduced by Sankai and colleagues in 2005, to refer to their exoskeleton \cite{Hayashi2005, Sankai2006}: it hints at the resemblance of soft wearable robots to clothing. }
Probably kindled by the unprecedented results of the Harvard exosuit (shown in Figure~\ref{Hardi_to_suit}.c), research and commercial interest in soft wearable robots have grown impressively fast and continue to do so. 

The present work provides a critical overview of the developments in the field of soft robotic suits. We start by proposing a definition of soft robotic suits and introduce a taxonomy to classify existing systems. We {\color{black} then review the state of the art} and highlight the core technologies that have grown alongside these robotic devices: (1) compliant actuation methods, (2) design principles for the human-suit interface and (3) sensors and strategies for intention-detection. We summarize the findings of the surprisingly high number of studies that evaluated the biomechanical effects of soft wearable robots on human motor performance. {\color{black}Throughout the manuscript, we comment on the key points of each section, highlight major open challenges and speculate on possible future directions.}

This survey has a focus on devices intended to assist the lower limbs, trunk and upper limbs, intentionally overlooking the advancements in soft wearable technologies assisting hand function. These make up a great part of the literature and have been nicely summarized in a recent review paper \cite{Chu2018}. {\color{black} Furthermore, we here discuss devices that purely make use of passive elastic elements to assist human movements. Although these are not technically robots \cite{InternationalOrganizationforStandardization}, they deserve to be mentioned: their technological development --- materials and means of attachment to the human body --- has recently accompanied that of their robotic counterparts. Furthermore, there are numerous documented instances of positive effects of passive devices on human movement, from which we could learn to design more efficient robotic devices.}

{\color{black} This manuscript does not intend to serve as a comprehensive list of all devices in the literature or on the market, but rather a means to highlight the main advancements of soft wearable robots from a critical standpoint.} By doing so, we hope to both help researchers that are new to the topic as they look for opportunities to uniquely contribute with their own work, and draw attention to areas within the field that are in need of further development.

\section{Terminology, definition and taxonomy}\label{section_taxonomy}
{ \color{black} The word ``exoskeleton'' comes from Greek \textgreek{έξω}, ``outer'' and \textgreek{σκελετός},  ``skeleton''. The latter part of the word was replaced with ``suit'' (e.g. \cite{Wehner2013}), to refer to the soft nature of the frame of the device. This term is now frequently used in the community, although the prefix ``exo'' is somewhat redundant for a suit.  Following the same etymological rationale, these robotic devices are  also referred to as ``exomuscles'' \cite{Simpson2020}, which more accurately reflects their working principle.

In this paper, we avoid using the oxymoron ``soft exoskeletons'' and use the general term ``soft robotic suits'' to encompass ``exosuits'', ``exomuscles'', and all devices that conform to Definition~\ref{soft_def}.}

\begin{definition}[]\label{soft_def}
Soft robotic suits are clothing-like robotic devices that wrap around a person's body and work in parallel with his/her muscles. Characteristic of soft robotic suits is that they rely on the structural integrity of the human body to transfer reaction forces between body segments, rather than having their own load-bearing frame. They thus act like an external layer of muscles rather than an external skeleton.
\end{definition}

 \begin{figure*}
    \centering
    \includegraphics[width = .7\textwidth]{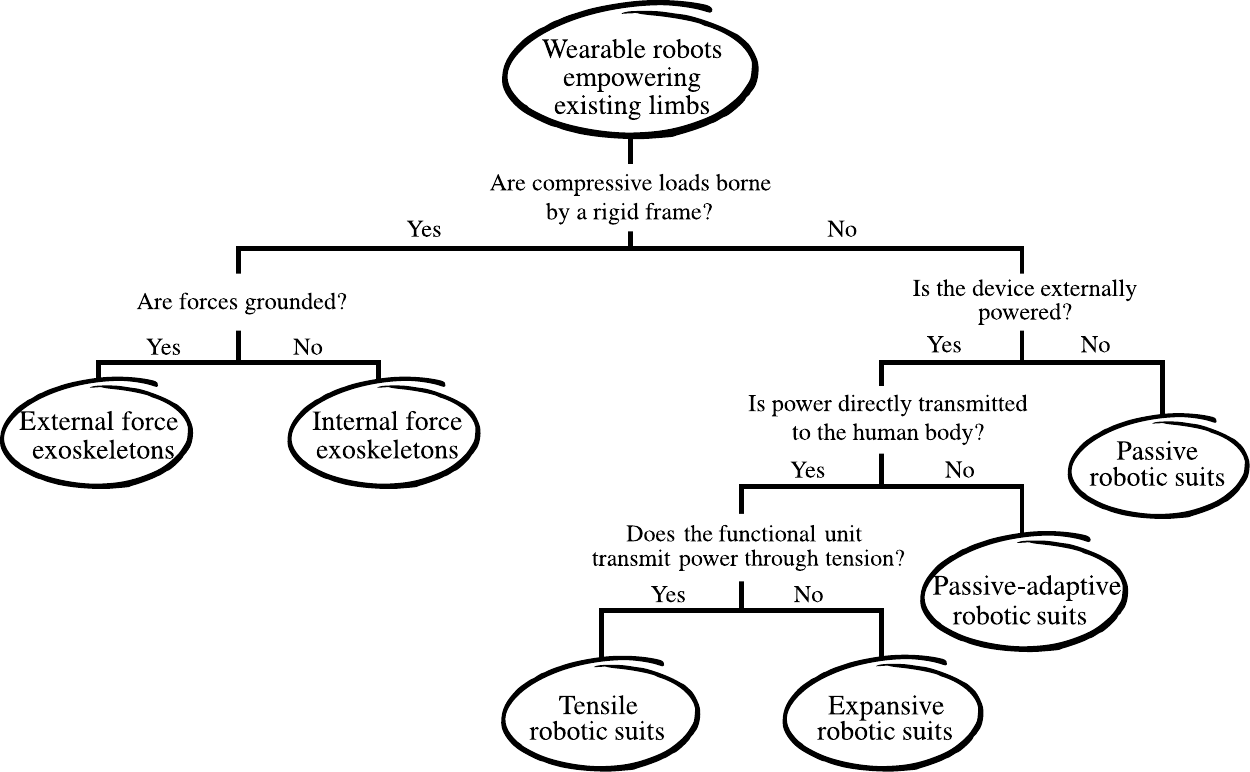}
    \caption{A taxonomy of soft robotic suits. The classification starts very broadly from wearable robots that assist existing limbs, as defined in \cite{Pons2008}. The first dichotomy identifies the difference between exoskeletons and soft robotic suits, the latter not having a rigid frame that bears compressing forces in stead of the human skeleton. Soft robotic suits are further categorized according to their source of power. Passive devices exploit gravitational energy, or kinematic energy from the human body, to selectively accumulate and release forces. Passive-adaptive devices exploit an external source of power to modulate their mechanical properties, thus having more versatility over the timing and magnitude of energy accumulation and release. Lastly, robotic suits that source power from an external supply and directly deliver it to the human body are divided in those that do so through a tensile element and those that do so through an expansive functional unit.}
    \label{taxonomy}
\end{figure*}

Our taxonomy of soft robotic suits (shown in Figure~\ref{taxonomy}) starts broadly from wearable robotic devices that supplement the function of an existing limb, versus those that substitute it, as defined in \cite{Pons2008}. Here, we propose a number of dichotomous keys, similar to those used in biology, to classify and distinguish the devices presented in the literature. The first dichotomy identifies the primary distinction between rigid exoskeletons and soft robotic suits: the latter do not have a rigid structure that bears compressing loads in parallel to the human skeleton. 

{\color{black} Note that this dichotomy is independent from the materials used to fabricate the device or from the level of compliance of the human-robot interface: the work from Tsagarakis and colleagues \cite{Tsagarakis2003}, for example, although using soft McKibben Pneumatic Artificial Muscles (PAMs) to actuate the device, would qualify as an exoskeleton because of the load-bearing frame in parallel with the human skeleton. Similarly, the device described in \cite{Jarrett2017}, in spite of the compliance attained through the soft elastomeric cores within its joints, falls under the class of exoskeletons.}

Exoskeletons can be classified in those that are grounded, also known as \textit{external force exoskeletons} or \textit{extenders}, and \textit{internal force exoskeletons} \cite{Pons2008}. The former transmit power to an external base, fixed or portable, and are typically used to extend human strength beyond its natural limits. Internal force exoskeletons are not grounded and the force transfer only occurs between the device and the user; a common application is complementing weak motor abilities. 

Soft robotic suits can be divided based on the source of the mechanical power they deliver to the body. Passive devices have no external means of power and rely on human movement or gravitational force to selectively store and release energy in different phases of movement. The Exoband, for example, does so through a pair of elastic bands that accumulate energy in the stance phase of walking, to support limb advancement during swing \cite{Panizzolo2019a}. 

An external power supply can be used to either directly power an actuator that generates assistive mechanical power or to alter the mechanical properties of the device. The latter case falls in the group of passive-adaptive soft robotic suits: the XoSoft, for instance, uses electromagnetic clutches to modulate the timing of engagement of passive elastic bands, allowing more control than a purely passive device over the magnitude and timing of energy storage and release \cite{DiNatali2019}. 

An active device, on the other hand, will transmit power from an external source to the human body directly. Means of actuation for active soft robotic suits can vary significantly and are the topic of Section~\ref{actuation}. A first broad dichotomy of actuation principles, however, is the distinction between devices that deliver power to a human joint by tensioning a functional unit, in a manner similar to skeletal muscles, and those that deliver power by expanding a functional unit.

The following two sections present an overview of the state of the art of soft robotic suits for human motion assistance. {\color{black} We have only included devices that conform to Definition~\ref{soft_def}, separated those assisting the upper and lower limbs and grouped them according to the taxonomy shown in Figure~\ref{taxonomy}. }

\section{Soft robotic suits for the lower Limbs}\label{lower_limbs}

\begin{table*}[htbp] 
	\renewcommand{\arraystretch}{1.2}
	\caption{\color{black} Soft robotic suits for the lower limbs.}
    \centering
    \begin{tabular}{l|c c c c c c }
    \hline
    \hline
    Device/Study  & \begin{tabular}{@{}c@{}} Year of \\ first publication \end{tabular} & Application & \begin{tabular}{@{}c@{}} Target \\ DoF(s) \end{tabular}   & Actuation &  Sensing & \begin{tabular}{@{}c@{}} Intention \\ detection \end{tabular}  \\ \hline
    \rowcolor{mygray}
    Pedomotor \cite{LeslieC.Kelley1919}& 1919 & n.a. & \begin{tabular}{@{}c@{}}Hip F/E, Knee F/E,\\ Ankle PF/DF \end{tabular}& \begin{tabular}{@{}c@{}}Steam engine, \\ tendons \end{tabular} & \begin{tabular}{@{}c@{}} Weighted lever, \\ spring \end{tabular}& Gait event \\
    Asbeck \textit{et al.} \cite{Asbeck2013} & 2013 & \begin{tabular}{@{}c@{}} Military \end{tabular} & Hip F, Ankle PF  & eMTUs &  \begin{tabular}{@{}c@{}} Load cells, \\ IMUs \end{tabular}& Gait event \\
    \rowcolor{mygray}
    Wehner \textit{et al.} \cite{Wehner2013}& 2013 & Military & \begin{tabular}{@{}c@{}} Hip F/E, Knee F/E\\  Ankle PF/DF \end{tabular}& PAMs  & Foot switch & Gait event \\
    Park \textit{et al.} \cite{Park2014}& 2014 & Healthcare &\begin{tabular}{@{}c@{}} Ankle PF/DF,\\  Ankle I/E \end{tabular} & PAMs &\begin{tabular}{@{}c@{}} IMUs, GRF\\  Strain sensors \end{tabular}   & ---\\
    \rowcolor{mygray}
    Park \textit{et al.} \cite{Park2014b}& 2014 & Healthcare & Knee F/E & PAMs  & --- &--- \\
    Superflex Inc. \cite{Cromie} & 2016 & Military & \begin{tabular}{@{}c@{}}   Ankle PF \end{tabular} & Twisted strings & --- & --- \\
    \rowcolor{mygray}
    \begin{tabular}{@{}l@{}} Awad \textit{et al.} \cite{Awad2017}  \end{tabular} & 2017 & Healthcare &Ankle PF/DF& eMTUs  & Load cells, IMUs &  Gait event \\
    Lee \textit{et al.}  \cite{Lee2017} & 2017 & Military &Hip E & eMTUs & Load cells, IMUs & Gait event \\
    \rowcolor{mygray}
    Myosuit, MyoSwiss \cite{Schmidt2017} & 2017 & Healthcare &  \begin{tabular}{@{}c@{}}   Hip E \\ Knee E \end{tabular} &\begin{tabular}{@{}c@{}}eMTUs, \\ Passive \end{tabular}  & IMUs  & \begin{tabular}{@{}c@{}} Gait event \end{tabular}  \\
    Sridar \textit{et al.} \cite{Sridar2017a}& 2017 & Healthcare  & Knee E & PIAs & IMUs, GRF & Gait event \\
    \rowcolor{mygray}
    ReStore\texttrademark\, Rewalk \cite{Awad2020a} & 2018 & Healthcare & Ankle F/E& eMTUs & IMUs & Gait event\\ 
    ExoBoot \cite{Chung2018} & 2018 & Healthcare & Ankle PF& PIAs & IMUs &Gait event \\
    \rowcolor{mygray}
    Thakur \textit{et al.} \cite{Thakur2018}& 2018 &\begin{tabular}{@{}c@{}} Healthcare \\ Industrial \end{tabular}& Hip F & PAMs  & GRF & Gait event  \\
    Thalman \textit{et al.} \cite{Thalman2019}& 2019 & Healthcare & Ankle PF/DF & PIAs + PAMs  & GRF & Gait event  \\
    \rowcolor{mygray}
    XoSoft $\beta$ \cite{DiNatali2019}& 2019  & Healthcare & Hip F, Knee F & \begin{tabular}{@{}c@{}} Passive-adaptive \end{tabular} & GRF & Gait event  \\
    Exoband \cite{Panizzolo2019a}& 2019  & Healthcare & Hip F& Passive & --- & ---  \\
    \rowcolor{mygray}
    Park \textit{et al.} \cite{Park2020}& 2020  & Healthcare & Hip Ab, Knee E & eMTUs & IMUs & Gait event\\
    Wang \textit{et al.} \cite{Wang2020}& 2020  & n.a.   & Ankle PF & eMTUs & IMUs & Breaststroke event  \\ 

    \hline
    \hline 
    \end{tabular}
    \begin{flushleft}
    \footnotesize{F =  flexion; E =  extension; Ab = abduction; PF = plantarflexion; DF =  dorsiflexion; eMTUs = electric motor-tendon unit; PIA = pneumatic interference actuator; PAM = pneumatic artificial muscle; IMU = inertial measurement units; GRF = ground reaction forces; n.a. = information not available; --- = the device did not feature the corresponding technology. }
    \end{flushleft}
    \label{table_low}
\end{table*}

{\color{black} Table~\ref{table_low} lists works that describe soft robotic suits for the lower limbs, in chronological order, highlighting their field of application and associated core technologies. The following sections discuss these works, grouping them according to the taxonomy outlined in Section~\ref{section_taxonomy}.}
 \begin{figure}[hbtp]
    \centering
    \includegraphics[width = .3\textwidth]{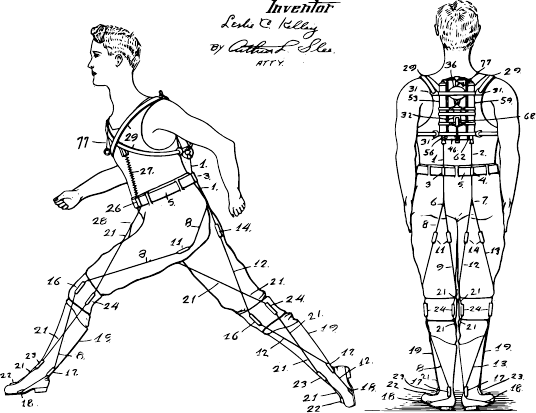}
    \caption{The Pedomotor, patented by  L. C. Kelly in 1919 \cite{LeslieC.Kelley1919}. The device used a set of wires, or artificial ligaments, arranged in parallel with the principal muscles responsible for running. Kelly even envisioned a way to control the timing of assistance, using the inertia of a weighted lever, carried on the backpack, that would open a valve upon heel-strike. Image adapted from \cite{LeslieC.Kelley1919}.}
    \label{Pedomotor}
\end{figure}
 \begin{figure*}[htbp]
    \centering
    \includegraphics[width = .8\textwidth]{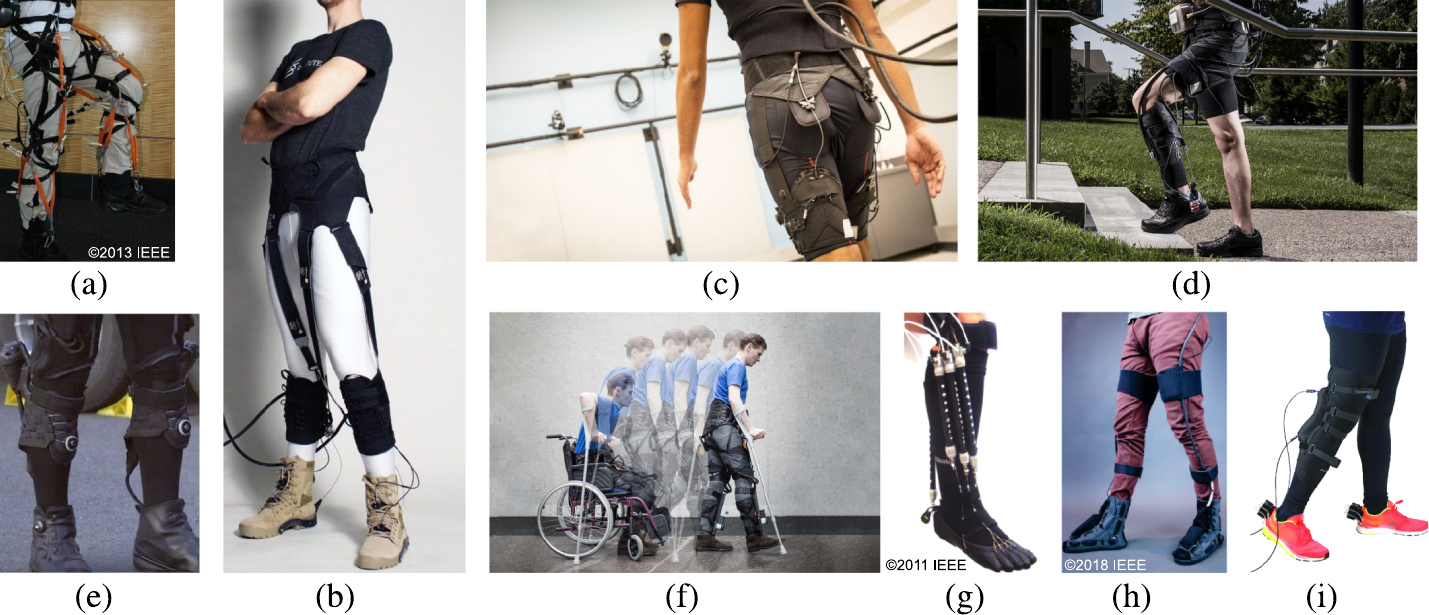}
    \caption{Soft robotic suits for the lower limbs, using tensile (a-g) and expansive (h-i) functional units. (a-d) The four  exosuit designs proposed by the Biodesign Lab at Harvard University \cite{Wehner2013,Asbeck2013,Lee2017, Awad2017}. Photo credits: (b) Wyss Institute at Harvard University/Harvard Biodesign Lab, (c) Wyss Institute at Harvard University/Harvard Biodesign Lab, (d) Rolex/Fred Merz. (e) The soft robotic suit for assistance of ankle plantar-flexion, developed by SRI International; the device features a high power density twisted string actuator \cite{Stevens2016}. Photo credit: Fusion TV. (f) MAXX, the device for walking assistance with a multi-articular tendon path \cite{Schmidt2017a}. Photo courtesy of NCCR Robotics. (g) McKibben PAMs were used for a bio-inspired design from Park and colleagues \cite{Park2011}, to support rehabilitation exercises of the ankle. The Exoboot (h) \cite{Chung2018} and the soft robotic suit designed by Sridar \etal{} \cite{Sridar2018} (i), instead, use expansive Pneumatic Interference Actuators (PIAs) to support plantar-flexion of the ankle and extension of the knee, respectively.}
    \label{exos_low}
\end{figure*}
\subsection{Tensile robotic suits}
What was likely the first technical design of a soft robotic suit, dating back to to 1919, was L. C. Kelly's ``Pedomotor'' \cite{LeslieC.Kelley1919}, a \textit{``...power device adapted to relieve the muscles utilized during the running operation of the anatomy from strain and fatigue.''} The device (Figure~\ref{Pedomotor}) consisted of two sets of artificial tendons per leg, routed along the lines of the principal muscles responsible for flexing and extending the hip, knee and ankle. Kelly envisioned that these tendons would be periodically pulled and released by a suitable motive power, carried around the shoulders and triggered by the impact of the user's feet with the ground.

Aside from the means of actuation (a steam-powered engine), Kelly had introduced some of the key concepts that, a century later, would make soft robotic suits so effective: a contractile network of tendons, working in parallel with the human muscles; a proximally-located actuation stage. Kelly's patent described with remarkable accuracy what we would now call a tensile soft roboti suit for the lower limbs. 

In 2013, the Biodesign Lab, led by C. J. Walsh, funded to bring forward the US Defense Advance Research Project Agency (DARPA) Warrior Web Bravo, published the design of two soft exosuits to assist locomotion \cite{Wehner2013,Asbeck2013}. Both systems were designed with the same goal and rationale to achieve it: to reduce the metabolic cost of walking of healthy individuals by applying short pulses of well-timed assistance, leveraging on a textile-based frame with negligible mechanical impedance and no kinematic restrictions. 

Both suits used tensile elements to assist locomotion but different actuators. The work from Wehner (Figure~\ref{exos_low}.a) consisted of a network of soft, inextensible webbing bands, actuated by McKibben PAMs \cite{Wehner2013}, arranged in an antagonistic manner to assist the hip, knee and ankle on the sagittal plane.  

The work from Asbeck \etal{}, instead, used DC electric motors and pulleys to reel in artificial tendons made of steel wire, routed from a backpack to the ankle joint through Bowden cables \cite{Asbeck2013}. 
The suit would assist ankle plantar-flexion and hip flexion during late terminal stance and pre-swing and be slack for the rest of the gait cycle. 

The work from Asbeck and colleagues highlighted two key advantages of soft robotic suits: (1) the robot could seamlessly transition from applying high forces to becoming  ``transparent'' (i.e. apply zero forces) to the wearer; (2) the Bowden cables allowed the tendon-driving unit to be positioned close to the center of mass of the suit's wearer, where the added mass has the least impact on the metabolic cost of walking \cite{Royer2005}. 

Over the subsequent seven years, the Biodesign Lab, using a very effective iterative design and testing procedure, refined the tendon-driven exosuit described in \cite{Asbeck2013} and made major contributions to the field, pushing the boundaries of how soft robotic suits can be used to both augment human abilities and assist people with neuromuscular impairments. We have identified three main exosuit designs for the lower limbs in their subsequent studies, based on the targeted joints and intended application: (1) a bi-articular exosuit for assisting hip flexion and ankle plantar-flexion, designed for augmenting healthy human walking (Figure~\ref{exos_low}.b); (2)  a mono-articular exosuit for assisting hip extension, designed for augmenting healthy human walking and running (Figure~\ref{exos_low}.c); (3) a mono-articular exosuit for assisting ankle plantar- and dorsi-flexion, designed to improve propulsive power and ground-clearance in patients with hemiparesis after stroke (Figure~\ref{exos_low}.d). A combination of (1) and (2) was presented in \cite{Panizzolo2016,Ding2017}. 

The multi-articular exosuit (Figure~\ref{exos_low}.b) used a bio-inspired strategy to actuate two joints with one motor-tendon unit, per leg. The muscles responsible for hip flexion and ankle plantar-flexion work in synergy between 20\% and 62\% of the gait cycle, to absorb energy during early to mid stance and actively assist push-off for limb advancement \cite{Eng1995}. 


The suit used a foot-switch to estimate an average gait period, adjusted every 5 steps, and triggered a force profile that reached its peak at approximately 50\% of the gait cycle \cite{Asbeck2015}. The foot switch was later augmented with a gyroscope MicroElectroMechanical Systems (MEMS) sensors, estimating the angular speed of the foot on the sagittal plane to allow for live walking cadance adaptation \cite{Asbeck2015a}. 

The exosuit in \cite{Asbeck2014} used the same rationale presented for the design of the multi-articular exosuit, but to assist hip extension (shown in Figure~\ref{exos_low}.c). One actuation module per leg, carried in a backpack, comprised of an electric motor driving a spool onto which a webbing ribbon was wound. The webbing ribbon was attached to the posterior side of a thigh brace and wound by the motor, upon heel-strike, to work in parallel with the hip extensor muscles during the loading response and early mid-stance phases of the gait cycle.
Heel-strikes were detected using foot-switches and the desired force profile was followed using a force-based position controller (detailed in Section~\ref{actuation}).
A more recent implementation of the exosuit for hip extension \cite{Kim2019} is completely mobile, with a total weight of \SI{5}{\kilo\gram}, 91\% of which is carried around the waist, and uses an online classification algorithm to seamlessly switch between walking and running assistance.

In \cite{Bae2015}, the Biodesign Lab proposed a unilateral soft exosuit to assist ankle plantar-flexion (PF) and dorsi-flexion (DF), with the goal of improving mobility in hemiparetic patients with residual motor function but inefficient walking patterns (suit shown in Figure~\ref{exos_low}.d). The suit described in \cite{Bae2015} addressed the need of improving ground-clearance in the swing phase of walking, with a DF module, while assisting propulsion at toe-off with a PF module. 
Assistance to the ankle was triggered by an online detection of events in the gait cycle, from a MEMS gyroscope placed on the foot. In \cite{Awad2017}, the authors present a mobile version of the suit that significantly improved economy and symmetry of walking in chronic stroke patients. 

The Stanford Research Institute Robotics group was also funded by the DARPA Warrior Web program to develop soft, comfortable and lightweight means of assisting human movements. Their solution was the Super-Flex (Figure~\ref{exos_low}.e), a soft robotic suit assisting ankle PF using a Twisted-String Actuator (TSA) (Section~\ref{actuation}) anchored on the back of the lower leg and pulling on a bootstrap on the heel.
The developers published the design of a high power-density actuator, called the FlexDrive \cite{Stevens2016}, used to power the suit, and a novel textile interface to distribute actuation forces evenly on the underlying soft tissues \cite{Witherspoon}. 

While most of the devices presented so far assisted human walking by providing propulsive forces before push-off, MAXX (Figure~\ref{exos_low}.f) \cite{Schmidt2017},  provided anti-gravity support in the stance phase of walking. MAXX used a tensile motor-tendon unit to transmit forces from a tendon-driving unit, placed on the shank, to the knee and hip. One artificial tendon per leg was routed from the tendon-driving unit to a bi-articular load path spanning the front of the knee and back of the hip.
Using two IMUs, placed on the trunk and shank, an incremental encoder on each of the motor's axis and a load cell in series with each artificial tendon, the device presented in \cite{Schmidt2017} delivered a force that increased with increasing bending of the knee joint. A ``transparency'' mode of operation allowed the user to turn off the assistance while keeping a slight pre-tension on the tendons. 
Haufe \etal{} later showed that a revision of the robot --- the Myosuit --- in transparency mode, does not affect healthy human kinematics \cite{Haufe2019} and, when active, is a safe and effective tool for training \cite{Haufe2020a, Haufe2020e}.  

Among the first to make use of McKibben PAM, Park and colleagues designed a rehabilitation device for the ankle joint \cite{Park2014} (Figure~\ref{exos_low}.g). The prototype described in \cite{Park2014} used a bio-inspired design approach, where four Mckibben PAMs were arranged so as to mimic the muscle-tendon-ligaments structure in the lower leg, to assist ankle PF, DF, inversion and eversion. The device used a redundant sensing strategy to estimate the position of the joint on the sagittal and frontal plane, consisting of two silicone-based strain sensors, and two IMUs. 

\subsection{Expansive robotic suits}
The Biodesign Lab at Harvard also investigated a radically different approach of assisting walking with a soft ``ExoBoot''  \cite{Chung2018}. The device (Figure~\ref{exos_low}.h) is a soft inflatable robotic boot that assists ankle plantar-flexion, without restricting other DoFs at the ankle, using a Pneumatic Interference Actuator (PIA). The pneumatic actuator relied on a cylindrical inflatable chamber, enclosed in a flexible but inextensible layer of fabric, to generate moments around the ankle, through a mechanism of self-intersection \cite{Nesler2018}. Using a low-level feedback pressure controller and IMUs to estimate gait events and time the assistance, the ExoBoot was shown to deliver up to \SI{29}{\newton\meter} in plantar-flexion, synchronized with the biological peak of torque production ($\approx$\SI{150}{\newton\meter}). 

PIAs were used in a similar fashion to assist extension of the knee joint during the swing phase of walking, in \cite{Sridar2020} (Figure~\ref{exos_low}.i). The inflatable actuator, enclosed in an inelastic fabric pocket, was held in position and the back of the knee with neoprene fabric and hook and loop straps. Using IMUs to detect gait events and a close-loop pressure controller, the authors delivered approximately \SI{25}{\%} of the biological torque required for knee extension.

The same research group, proposed an interesting approach for delivering assistive forces at the ankle joint, using PIAs \cite{Thalman2019} in combination with tensile PAMs. A contractile inflatable actuator, anchored on the anterior of the lower leg and on the dorsal foot, dorsi-flexed the foot during the swing phase of walking, while a pair of PIAs, attached to the back of the foot, modulated the stiffness of the ankle joint upon heel strike. The device was controlled using two pairs of force sensitive resistors to detect heel strike and toe-off, and fluidic pressure sensors to close feedback control loops on the pressure in the PAMs. 

\subsection{Passive-adaptive robotic suits}
The EU-funded XoSoft project resulted in the development of a quasi-passive soft robotic suit for walking assistance \cite{DiNatali2019}. The device used two passive elastic elements, traversing the  hip anteriorly and the knee posteriorly, connected in series with electromagnetic clutches, placed in a unit carried as a backpack. Engagement and disengagement of the clutches during the gait cycle allowed to selectively activate or deactivate the energy storage and release properties of the elastic elements. The authors used a pair of instrumented insoles to detect gait events and control the timing of engagement/disengagement of the clutches.  A single-case study on a chronic stroke patient showed that the XoSoft improved ground clearance and gait symmetry \cite{DiNatali2019}. 

\subsection{Passive robotic suits}
With a total weight of \SI{645}{\gram}, the ExoBand is, to our knowledge, the lightest assistive suit designed so far \cite{Panizzolo2019a}. The passive device comprises of one waist and two thigh braces, used to anchor two elastic bands, traversing the hip joint anteriorly. The anteriorly-placed springs store energy in the phase preceding maximum hip extension while supporting leg deceleration, and return that energy in the swing phase, to support limb advancement. The authors showed that, despite not delivering external power to the human body, the Exoband could reduce the metabolic cost of walking at \SI{1.1}{\meter/\second} by an average 3.3\%, compared to unassisted walking, in a sample of nine older adults. The biomechanical effects of a similar device were thoroughly analysed in \cite{Haufe2020c}.

\section{Soft robotic suits for the upper limbs}\label{upper_limbs}
\begin{table*}[htbp]
	\renewcommand{\arraystretch}{1.2}
	    \caption{\color{black} Soft robotic suits for the upper limbs and trunk.}
    \centering
    \begin{tabular}{l|c c c c c c }
    \hline
    \hline
    Device/Study  & \begin{tabular}{@{}c@{}} Year of \\ first publication \end{tabular} & Application & \begin{tabular}{@{}c@{}} Target \\ DoF(s) \end{tabular}   & Actuation  &  Sensing & \begin{tabular}{@{}c@{}} Intention \\ detection \end{tabular}  \\ \hline
    \rowcolor{mygray}
    Kobayashi \textit{et al.} \cite{Kobayashi2004} & 2004 & n.a.  & \begin{tabular}{@{}c@{}}  Shoulder F, Ab \end{tabular} & PAMs &  --- & ---  \\
    Ueda \textit{et al.} \cite{Ueda2007} & 2007 & n.a.  & \begin{tabular}{@{}c@{}}  Elbow F/E  \end{tabular} & PAMs &  --- & ---  \\
    \rowcolor{mygray}
    Koo \textit{et al.} \cite{Koo2014} & 2014 & Healthcare & \begin{tabular}{@{}c@{}}   Shoulder F, Ab \\ Elbow F \end{tabular}& eMTUs +PIAs & Reed switches & Thimble movement \\
    Cappello \textit{et al.} \cite{Cappello2016}& 2016 & Healthcare & Elbow F/E & eMTU & --- & --- \\ 
    \rowcolor{mygray}
    Koh \textit{et al.} \cite{Koh2017} & 2017 & Healthcare & Elbow F/E & PIAs  & EMG electrodes & Muscular activity \\
    Exomuscle \cite{Simpson2017a} & 2017 & Healthcare &  Shoulder Ab & PIAs &  --- & --- \\
    \rowcolor{mygray}
    Auxilio \cite{Gaponov2017} & 2017 & Healthcare & \begin{tabular}{@{}c@{}}  Shoulder F, Ab \\ Elbow F \end{tabular}  & \begin{tabular}{@{}c@{}}  Twisted strings \end{tabular}  & --- & --- \\
    O'Neill \textit{et al.} \cite{ONeill2017} & 2017 & Healthcare & \begin{tabular}{@{}c@{}}   Shoulder Ab, HF/HE \end{tabular}& PIAs & --- & ---  \\
    \rowcolor{mygray}
    Park \textit{et al.} \cite{Park2017a} & 2017 & Healthcare & \begin{tabular}{@{}c@{}}   Shoulder F, Ab \end{tabular}& Passive & --- & --- \\
    Xiloyannis \textit{et al.} \cite{Xiloyannis2019}& 2018 & \begin{tabular}{@{}c@{}}  Healthcare, \\ Industrial \end{tabular}& Elbow F/E& eMTU & \begin{tabular}{@{}c@{}} Load cells,\\ Stretch sensor \end{tabular} & Interaction force  \\ 
    \rowcolor{mygray}
    Kim  \textit{et al.} \cite{Kim2018} & 2018 & Industrial & \begin{tabular}{@{}c@{}}   Shoulder F\\ Elbow F \end{tabular} & eMTUs & Microphone & Speech recognition  \\
    Li \textit{et al.} \cite{Li2018} & 2018 & Healthcare & \begin{tabular}{@{}c@{}} Shoulder F, Ab/Ad, \\ Elbow F/E, \\  Wrist F/E Ad/Ad \end{tabular}  & eMTUs &  Load cell & --- \\
    \rowcolor{mygray}
    CRUX \cite{Lessard2018} & 2018 & Healthcare &\begin{tabular}{@{}c@{}} Shoulder F/E, Ab/Ad, \\ Elbow F/E, \\ Wrist P/S\end{tabular} & eMTUs &  IMUs & Arm mirroring  \\
    Sridar \textit{et al.} \cite{Sridar2018b} & 2018 & Healthcare & Shoulder F & PAMs & IMUs & Event detection \\

    \rowcolor{mygray}
    Nguyen \textit{et al.} \cite{Nguyen2019} & 2019 & Healthcare & Elbow F & PIA  & --- & ---  \\
    Abe \textit{et al.} \cite{Abe2019} & 2019 & Healthcare &\begin{tabular}{@{}c@{}}  Shoulder F/E, HF/HE \\   Elbow F/E \end{tabular} & PAMs  & --- & --- \\
    \rowcolor{mygray}
    SMA suit \cite{Park2019} & 2019 & \begin{tabular}{@{}c@{}}  Healthcare, \\ Industrial \end{tabular} &  Elbow F & SMA & --- & --- \\
    Hosseini \textit{et al.} \cite{Hosseini2020}& 2019 &  Industrial & \begin{tabular}{@{}c@{}} Elbow F \\ Shoulder F \end{tabular} & Twisted strings & \begin{tabular}{@{}c@{}} Force sensors, \\ EMG electrodes \end{tabular}& Muscular activity \\
    \hline
    Trunk:  & & & & & &  \\
    \hline
    \rowcolor{mygray}    
    Govin \textit{et al.} \cite{Govin2018} & 2018  & Healthcare & Spinal E & PIAs  & IMUs & Trunk position \\
    Lamers \textit{et al.} \cite{Lamers2018} & 2018  & Industrial & Spinal E & Passive  & --- & --- \\
    \rowcolor{mygray}
    Liftsuit, Auxivo AG & 2020  & Industrial & Spinal E & Passive  & --- & --- \\%
    SoftExo, Hunic & 2020  & Industrial & Spinal E & Passive  & --- & --- \\%
    \hline
    \hline
    \end{tabular}
  \newline
    \begin{flushleft}
    \footnotesize{ F =  flexion; E =  extension; Ab = abduction; Ad = adduction; HF =  horizontal flexion; HE =  horizontal extension; P = pronation; S = supination;  eMTU = electric motor-tendon unit; PIA = pneumatic interference actuator; PAM = pneumatic artificial muscle; IMU = inertial measurement units; EMG = Electromyography; n.a. = information not available; --- =  the device did not feature the corresponding technology. }
    \end{flushleft}
    \label{table_up}
\end{table*}
 \begin{figure*}
    \centering
    \includegraphics[width = .8\textwidth]{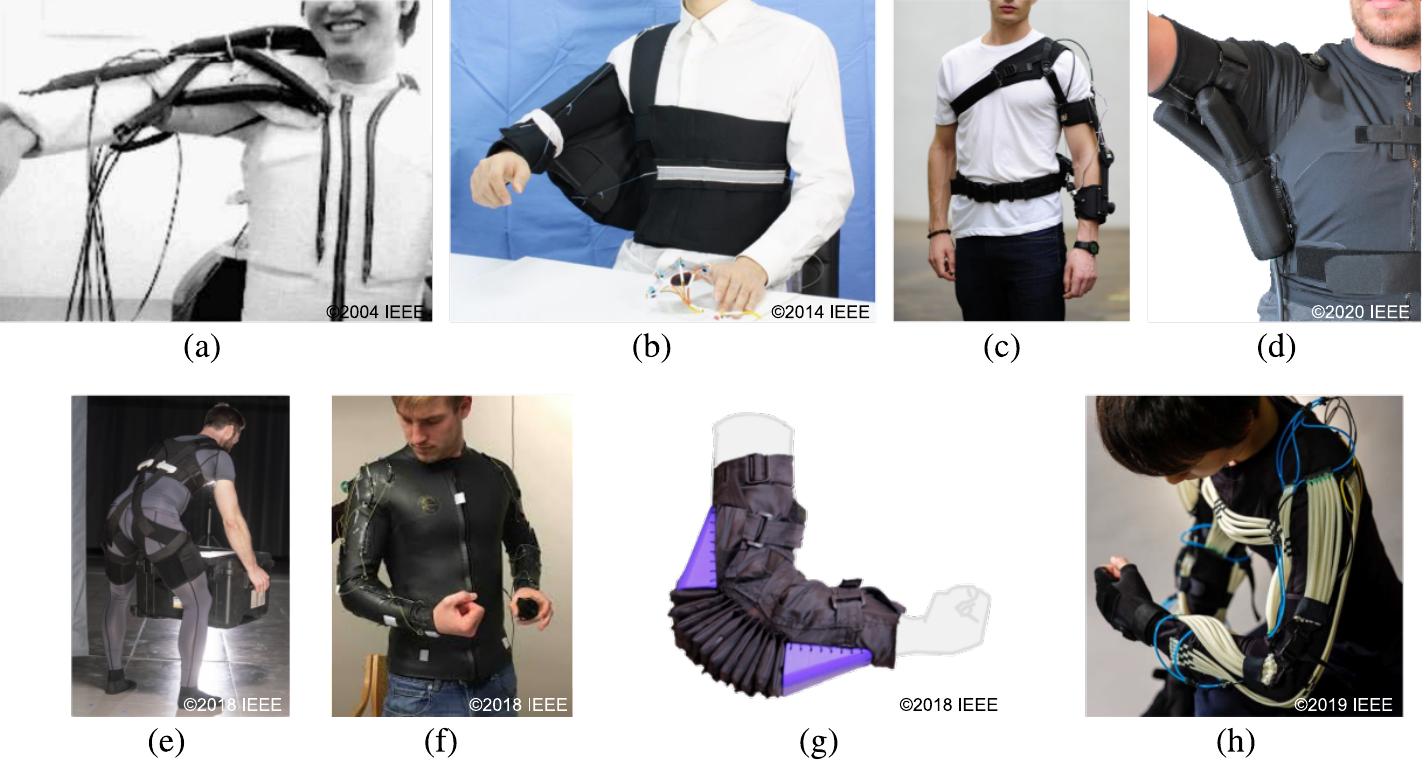}
    \caption{Soft robotic suits for the upper limbs. (a) One of the very first prototypes of a soft robotic suit for the upper limbs was proposed by Kobayashi; the ``muscle suit'' used McKibben PAMs to support the shoulder against gravity \cite{Kobayashi2004}. (b) The device described in \cite{Koo2014} supported a bi-articular feeding movement through a combination of artificial tendons and a passive inflatable chamber under the axilla. (c) The tendon-driven robotic device in \cite{Xiloyannis2019}, for  supporting the elbow joint against gravity. (d) The device presented in \cite{ONeill2020}, to support shoulder abduction, using an inflatable PIA. (e) One of prototypes that resulted in the HeroWear Apex suit, employing two elastic bands to support the lumbar spine \cite{Lamers2018}. (f) The CRUX robot uses a combination of soft and hard components to distribute forces homogeneously on a base layer of neoprene \cite{Lessard2018}. (g) The soft robotic suit proposed in \cite{Nguyen2019} uses a variation of a PIA consisting of a series of adjacent inflatable chambers. (h)  A unique approach is that presented in \cite{Abe2019}, consisting of a miniature McKibben actuators, arranged in parallel to achieve higher forces or in series for longer strokes. }
    \label{Upper_limbs}
\end{figure*}

{\color{black} Table~\ref{table_up} lists works that describe soft robotic suits for the upper limbs and trunk, in chronological order, highlighting their field of application and associated core technologies. The following sections discuss these works, grouping them according to the taxonomy outlined in Section~\ref{section_taxonomy}.}

\subsection{Tensile robotic suits}
In early 2004, Kobayashi \etal{}  presented the design of  the ``muscle suit'' \cite{Kobayashi2004} (Figure~\ref{Upper_limbs}.a), consisting of a network of McKibben PAMs, anchored to a semi-rigid, joint-less garment made of fabric and urethane. Controlling the pressure of the PAMs, Kobayashi verified the feasibility of this concept to assist shoulder abduction.

The research group led by Prof. Ogasawara later developed a similar tensile, McKibben-powered exomuscle \cite{Ueda2007}. Fully embracing the idea of the robot as a layer of external muscles, Ueda \etal{} used a bio-inspired topology of the actuators for the elbow and wrist: they arranged the McKibben muscles so that they ran in parallel to their biological counterparts. 

In 2014, Koo and colleagues presented a soft robotic suit designed to assist the movement of bringing food to the mouth \cite{Koo2014}. The device (Figure~\ref {Upper_limbs}.b), specifically targeting people with muscular weakness caused by polyomyositis, used a combination of off-board motor-tendon units and an inflated chamber to support the shoulder and elbow. 
The device was later augmented with a passive mechanism, working in parallel with the motor, to reduce power-consumption of the system \cite{Park2015}. The passive component alone was shown to provide sufficient assistance to delay the onset of muscular fatigue in healthy participants \cite{Park2017a}.

The group led by L. Masia designed a tensile soft robotic suit for assisting the elbow joint \cite{Cappello2016}. The suit was actuated by an electric motor, driving a pulley around which two dyneema cables were routed in opposite directions, so as to work in an agonist-antagonist fashion. 
The device was further revised (Figure~\ref{Upper_limbs}.c), with an on-board, mobile actuation scheme \cite{Chiaradia2018} and shown to reduce muscular activity and delay the onset of fatigue in healthy users \cite{Xiloyannis2019}.

 Kim \etal{} used the same principle to support industrial tasks. The soft robotic suit in \cite{Kim2020}  used  a combination of soft and rigid materials and a Bowden cable transmission, to support bi-manual shoulder elevation and elbow flexion during lifting operations.

The CRUX (Figure~\ref{Upper_limbs}.f) exomuscle was designed to support stroke survivors using a set of six DC motor-tendon units to assist the shoulder, the elbow and the wrist joints. Characteristic of the device was the idea of designing the suit's load-paths using the concept of tensegrity, i.e. a mixture of soft and rigid elements, combined so that the forces may be distributed homogeneously on a base of neoprene fabric. 
In a case-study with impaired users \cite{Lessard2018}, the authors reported a reduction in the heart rates of participants using the device during a bicep contraction task.

More recently, the idea of network of pneumatic contractile elements that work in parallel with the human muscles, introduced in \cite{Kobayashi2004} was implemented by \cite{Abe2019}, in a more granular form (Figure~\ref{Upper_limbs}.h): a network of  miniature McKibben actuators were arranged in parallel to achieve higher forces or in series for longer strokes.

A  different approach was used by Park and colleagues, in \cite{Park2019}. The authors describe a soft robotic suit to assist flexion of the elbow using a shape-memory-alloy-based muscle. The muscle consists of coil springs fabricated from a Nickel-Titanium (NiTi) wire, arranged in parallel to generate peak forces of \SI{120}{\newton}, with a weight of only \SI{24}{\gram} and a transition temperature of \SI{40}{\degree}.

\subsection{Expansive robotic suits}
Building on their extensive knowledge on wearable soft robots, the Biodesign Lab at Harvard designed a soft robotic suit to support the shoulder against gravity. O'Neill and colleagues' device features three inflatable PIA: one supported abduction of the shoulder and the remaining two worked in an antagonist fashion to apply torques on the horizontal plane \cite{ONeill2017}.  Characteristic of the device was a layer of flexible elements, mimicking the arrangement of the cruciate ligaments, to hold in place the actuators responsible for abduction.

A re-iteration of the device \cite{ONeill2020}, featured a single Y-shaped inflatable PIA that cradled the anterior and posterior of the arm, stably supporting it against gravity (Figure~\ref{Upper_limbs}.d). The new design could deliver up to \SI{16}{\newton\per\meter} at an abduction angle of \SI{90}{\degree} and input pressure of \SI{136}{\kilo\pascal}. The suit, donned by patients with severe stroke, was shown to both improve range of motion of the patient and reduce fatigue of the therapist assisting the patient.  

A bending moment can be similarly created by a network of interfering PIAs. In \cite{Nguyen2019}, the authors use soft cylindrical actuators, arranged in an array and constrained along the base with inextensible fabric, to achieve predictable bending moments and a linear torque-pressure characteristic. When inflated, the individual actuators interact with one another, to produce a bending motion about the elbow joint. The authors demonstrated the effectiveness of this topology in assisting flexion of the elbow joint (Figure~\ref{Upper_limbs}.g)

\subsection{Passive robotic suits}
Lamers and colleagues designed an elastic suit that works in parallel with the erector spinae group to support extension of the lumbar spine \cite{Lamers2018}. The device was made of a shirt and a pair of shorts, connected posteriorly by two crossing elastic bands with a stiffness of $\approx$\SI{800}{\newton\meter} (Figure~\ref{Upper_limbs}.e). To our knowledge, a revision of the device is now on the market, under the name of Apex (HeroWear).
Similar designs was recently commercialized by Hunic GmbH, with the SoftExo suit, and by the ETH spin-off Auxivo AG, with the LifSuit.

\section{Core technologies}
The rapid rise of soft robotic suits was strongly contingent on the development of enabling core technologies. This section focuses on the fundamental technical advancements that soft robotic suits exploit to assist human movements. We identify and detail three core technologies: (1) actuation paradigms and their low-level controllers;  (2) design of the physical human-robot interface (pHRI) and (3) sensing and control strategies for intention-detection. 

\subsection{Actuation and low-level control}\label{actuation}

\begin{figure*}[htbp]
    \centering
    \includegraphics[width = .75\textwidth]{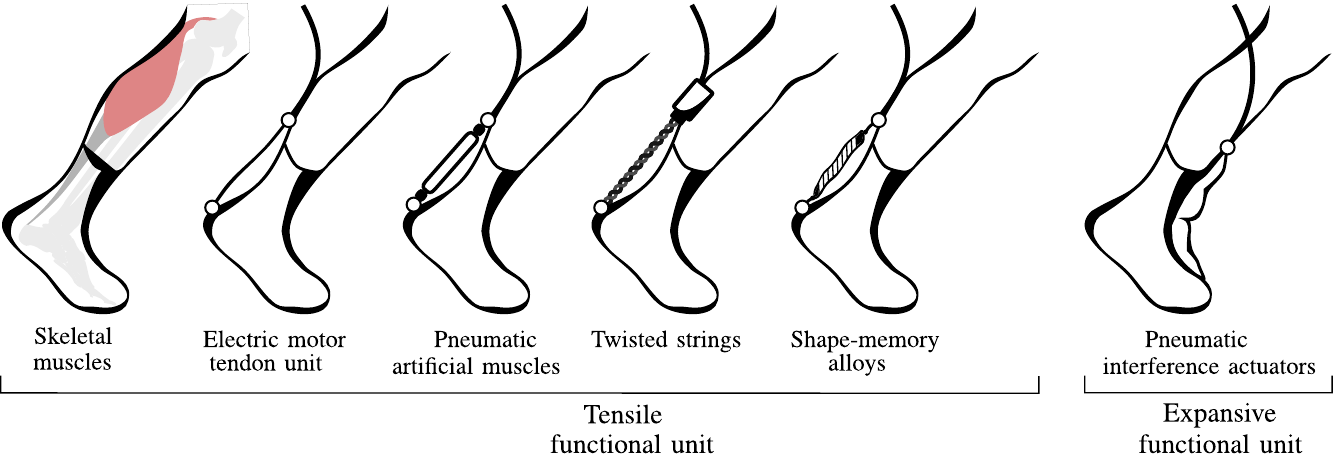}
    \caption{Modes of actuation of soft wearable robots. Skeletal muscles transmit forces to the joints through a muscle-tendon unit, capable of working only in tension. Forces are generated by the muscles, typically located away from the joint, and transmitted to the skeleton by tendons. Tensile soft robotic suits work in a similar manner, shortening a contractile element to apply a tensile force on the skeleton. Electric motor-tendon units comprise an electric motor, located away from the joint, that winds and unwinds an artificial tendon.  PAMs, instead, generate tensile forces by inflation of an elastic bladder, enclosed in an inextensible weave. Twisted string actuators convert rotary to linear power by twisting two or more  artificial tendons around a common axis, parallel to their length. Shape-memory alloy wires benefit from a very high force-weight ratio but have limited contraction and relaxation velocities. Pneumatic Interference Actuators (PIAs) work by self-intersection of a fabric bladder, arising from a change in its geometry.}
    \label{Contractile_expansive}
\end{figure*}

When classifying soft robotic suits, we divided active devices into those actuated by tensile or expansive functional units. This dichotomy is further elaborated in Figure~\ref{Contractile_expansive}: tensile soft robotic suits transmit positive power to the human joints by tensioning a functional unit, in a manner very similar to that of skeletal muscles. Those found in literature so far contract through four distinct mechanisms: (1) using an electric motor that wraps an artificial tendon around a pulley, (2) using McKibben PAMs, (3) twisted strings actuators or (4) shape-memory alloys. Expansive soft robotic suits, instead, assist human movements by expanding a folded, self-intersecting bladder filled with a fluid; these actuators are commonly referred to as PIAs \cite{Nesler2018}.

{\color{black} While soft artificial muscles rely on a large variety of principles and actuation mechanisms \cite{Zhang2019a}, only a subset of these comply to the requirements of bandwidth, safety and power density required to support human movements. The following sections elaborate on the functioning principles, advantages and shortcomings of each of the actuation solutions currently used for soft robotic suits.}

\subsubsection{Electric motor-tendon unit}
It is said of the 16$^{th}$ century polymath Leonardo Da Vinci, that \textit{``when he dissected a limb and drew its muscles and sinews, it led him to also sketch ropes and levers''}\cite{Isaacson2017}. Indeed, skeletal muscles in our body work much like an actuator-tendon couple.  { \color{black} One of the reasons why this configuration has survived the pitiless selection of nature is that it allows for remote actuation: the actuator can be located away from the actuated joints}. This reduces the mass of moving components and allows for a more efficient and responsive transmission.  

Three factors have contributed to make this the most adopted mechanisms for powering active soft robotic suits, so far: (1) artificial tendons can be routed between two distant points, in flexible and lightweight Bowden cable sheaths; (2) there is a well-established body of knowledge and technology available for controlling electric motors; (3) cables can be slacked, seamlessly turning the suit into a passive garment that does not affect human movements. This is particularly important during walking, if the device needs to deliver forces only in specific phases of the gait cycle. 

The downsides of tendon-driven soft robotic suits are (1) high shear forces on the skin and (2) low mechanical efficiency. The former is common to all contractile actuators. The latter is a well-known characteristic of Bowden cable transmissions, arising from the combination of losses due to friction between the tendon and its sleeve and compliance of the Bowden sheaths. Other than introducing the need for large power supplies, these phenomena complicate the control requirements: stick-slip and backlash effects deteriorate accuracy, bandwidth and stability of force-tracking \cite{Schiele2006}.

Nevertheless, there is a wide body of literature on how to achieve accurate control of position, velocity and force delivered with electric motor-tendon actuators. In \cite{Dinh2017}, the authors propose to control the force delivered by a soft, tendon-driven soft robotic suit using a hierarchical controller that compensates for friction and backlash in the transmission using an adaptive paradigm. 

\begin{figure}[htbp]
    \centering
    \includegraphics[width = .35\textwidth]{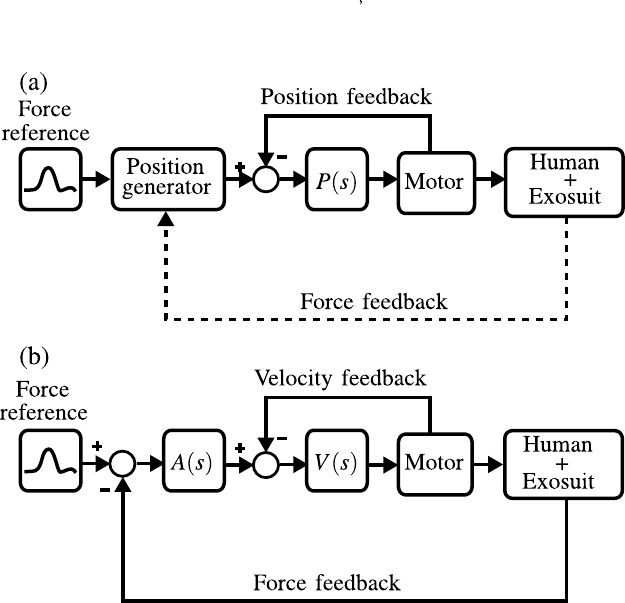}
    \caption{Force-tracking control paradigms used in soft tendon-driven robotic suits. (a) Force-based position control, used for the Harvard exosuit, e.g. in \cite{Quinlivan2017}. The position generator maps a reference force to a displacement profile of the tendon, followed by a position controller $P(s)$. Force feedback (dashed line) is used to update the displacement profile, after each step. (b) Admittance-based force tracking \cite{Lee2017b,Xiloyannis2019}. A virtual admittance $A(s)$ maps a reference force to a desired velocity of the tendon, followed by a high-gain velocity controller $V(s)$, closed on the motor axis.}
    \label{Controllers_electric}
\end{figure}

A more pragmatic approach is to use a high-gain position or velocity loop, cascaded with the force controller. Figure~\ref{Controllers_electric}.a shows a simplified schematics of a force-based position feedback controller used to track a desired assistive profile, as described in \cite{Quinlivan2017}. This paradigm makes use of an experimentally-identified human-suit series stiffness \cite{Asbeck2015}, to map a force reference to a desired position of the tendon. This desired position is tracked by a feedback controller, $P(s)$, closed on the motor's encoder. The outer force loop is ``loosely'' closed (dashed line) on a load cell in series with the tendon: the force feedback measurement is used to update the cable displacement, only once every step, to maintain a consistent force application.

Figure~\ref{Controllers_electric}.b shows an alternative approach for indirect force control, comprising of an inner velocity and outer force control loops. Here, $A(s)$ is a desired virtual admittance, mapping the desired force profile to a desired velocity, followed by a cascaded high-gain velocity controller $V(s)$. This paradigm, complemented with a feedforward component, was used in \cite{Lee2017b} to improve the tracking accuracy and bandwidth of the Harvard hip exosuit. A similar approach was used to provide anti-gravity support using a tendon-driven exosuit for the upper limbs \cite{Xiloyannis2019} and shown to outperform direct force control paradigms in a device for supporting shoulder elevation \cite{Georgarakis2020}.

{\color{black}To our knowledge, there are no implementations of impedance control paradigms for force-tracking in soft wearable robots. The reason probably lies in the notorious instability of the inner force loop of an impedance controller when one wants to render a stiff behaviour, i.e. accurately track a force profile. The simplest workaround this problem --- to substitute the inner force control with a current controller, effectively implementing the impedance controller proposed by Hogan \cite{Hogan1985} --- would require a very efficient and back-drivable transmission. }

\subsubsection{PAMs}
PAMs are soft fluidic tensile actuators that deliver mechanical power through compression of air. The most common PAMs are McKibben actuators, used, for example, for one of the first prototypes of the Harvard exosuit \cite{Wehner2013} and the ankle device from Park and colleagues \cite{Park2014}.  McKibben actuators are made of an elastic airtight bladder wrapped within helically-woven inextensible fibers. Upon inflation, the bladder loads the helical braid, expanding radially while contracting along the length of the muscle, thus producing an axial force.

PAMs are an appealing solution for soft robotic suits for their high power-to-weight ratio, muscle-like force-length properties and intrinsic compliance. Commercially-available McKibben actuators have strokes of up to 25\% of their resting length and are intrinsically safe thanks to their decreasing force characteristics for increasing contraction length \cite{Chou1996}. Substituting air with a liquid would greatly improve bandwidth and contraction forces but lead to an unacceptable increase in weight for wearable applications \cite{Mori2010}. 
 
While PAM actuators have a high power-to-weight ratio, the compressors or sources of compressed air required to power them are still heavy and bulky, limiting their portability. Chemical reaction-based pneumatic energy may be a promising path: hydrogen peroxide could be a good candidate for a more energy-dense source of pressure \cite{Onal2017}.

Due to the non-linear behaviour of PAMs, most of the soft robotic suits driven by PAMs in the literature use practical but limited control approaches such as bang-bang controllers \cite{Thalman2019,Wehner2013,Kobayashi2004}, feedforward position control \cite{Park2011}, proportional feedback position control \cite{Park2011} or PID pressure regulation \cite{Abe2018}. While effective, these control paradigms do not allow for the fine level of tuning of assistive profiles that can be achieved with electric motor-tendon actuators.
The development of a practical and effective method to control position and force of PAMs would greatly improve their adoption. 
\begin{table*}[htbp]%
	\renewcommand{\arraystretch}{1}
	\caption{Characteristics of skeletal muscles and actuation methods for soft robotic suits}
    \resizebox{\textwidth}{!}{
    \small
    \begin{tabular}{l| c c c c c c c }
    \hline
    \hline
     & \begin{tabular}{@{}c@{}} Power density \\(\SI{}{\watt\per\gram}) \end{tabular}  &  {\color{black}\begin{tabular}{@{}c@{}} Force density \\ (\SI{}{\newton\per\gram}) \end{tabular} }&  \begin{tabular}{@{}c@{}} Bandwidth\\ (\SI{}{\hertz}) \end{tabular} &  \begin{tabular}{@{}c@{}}Efficiency \\ (-) \end{tabular}  & Advantages  & Disadvantages & {\color{black} Examples} \\ \hline
    \begin{tabular}{@{}c@{}} Skeletal muscles \end{tabular} \cite{Pennycuick1984} & \begin{tabular}{@{}c@{}} 0.4 \\ \end{tabular}  & n.a. & n.a. & 0.35 & n.a. & n.a. & n.a. \\ 
    \rowcolor{mygray}
    \begin{tabular}{@{}l@{}} Electric motor-tendon \\ unit  \end{tabular} & 0.6-0.7 & 0.6-0.8 & 20$^{\dagger}$  & 0.6 & \begin{tabular}{@{}c@{}} Linearity \\ Bandwidth \end{tabular} &\begin{tabular}{@{}c@{}} High  shear forces \end{tabular} & \cite{Kim2019, Xiloyannis2019} \\
    \begin{tabular}{@{}l@{}} Pneumatic artificial  \\ muscles \cite{Beyl2014} \end{tabular} & 10$^{*}$& 10-14$^{*\top}$ & 6$^{\ddagger}$ & 0.3 & \begin{tabular}{@{}c@{}} Compliance \end{tabular} & \begin{tabular}{@{}c@{}} Pressure source \\ Non-linearity  \\ Short stroke\end{tabular} & \cite{Wehner2013, Kobayashi2004}\\ 
    \rowcolor{mygray}
    \begin{tabular}{@{}l@{}} Twisted strings \\actuators \cite{Palli2013} \end{tabular}& 0.8 & 2-2.5 & 2-3$^{\mathsection}$  & 0.7 & \begin{tabular}{@{}c@{}}High force density \end{tabular} & \begin{tabular}{@{}c@{}} Short lifetime  \\ Low bandwidth\end{tabular} & \cite{Stevens2016, Hosseini2020}\\ 
    \begin{tabular}{@{}l@{}} Shape-memory \\ alloys \cite{MohdJani2014a} \end{tabular}& 50 & 4-5 & 1-2$^{\mathparagraph}$  & 0.13 & \begin{tabular}{@{}c@{}} High power  density \end{tabular} & \begin{tabular}{@{}c@{}} Low bandwidth  \\ Short stroke \\ Low efficiency \end{tabular} & \cite{Park2019, Jeong2019}\\ 
    \rowcolor{mygray}
    \begin{tabular}{@{}l@{}} Pneumatic interference \\ actuators \cite{Nesler2018} \end{tabular} & 3.5-7.2$^{*}$& 0.17$^{*\triangleleft}$ & n.a. & n.a. & \begin{tabular}{@{}c@{}} Compliance \\ Comfort \end{tabular} & \begin{tabular}{@{}c@{}} Pressure source \\ Non-linearity \end{tabular} & \cite{ONeill2020a, Nguyen2019}\\ 
    \hline
    \hline
    \end{tabular}
    }
    \begin{flushleft}
    \footnotesize{$^*$Not including the compressor. $^{\top}$ Data from the Festo Fluidic Muscle DMSP. $^{\triangleleft}$ Torque-density in \SI{}{\newton\meter\per\gram}, from \cite{Veale2020}. $\dagger$Close-loop force control, peak-to-peak amplitude \SI{200}{\newton}. $^{\ddagger}$Close-loop torque control, peak-to-peak amplitude \SI{50}{\newton\meter}. $^{\mathsection}$Close-loop force control, peak-to-peak amplitude \SI{20}{\newton}. $^{\mathparagraph}$Close-loop force control, peak-to-peak amplitude \SI{30}{\newton}. n.a. = information not available.}
    \end{flushleft}
    \label{table_actuation}
\end{table*}
\subsubsection{Twisted strings}
Twisted string actuators (TSA) are an interesting variation to the classic tendon-driven system. Instead of tensioning a wire by wrapping it around a pulley, TSA work by twisting two or more parallel tendons, co-axial to the motor, about each other. Their first applications go back to machines like catapults and ballistae \cite{May2010}, where they were used for the high mechanical advantage they gave to the operators.

TSA have a few features that make them ideal for powering soft robotic suits. They can achieve high linear forces with low overall weight, thanks to the intrinsic high reduction of the twisted configuration and conversion of rotary to linear motion without additional bulky components (e.g. ball screws or gears). Moreover, the efficiency of a twisted string transmission can be as high as 92\%.  An example of these advantages is the actuator developed by Stevens \etal{}, to power the SuperFlex suit, briefly described in \cite{Stevens2016} and shown in Figure~\ref{exos_low}.f.

Twisted-string actuators present a few unique challenges. Notably, they exhibit a nonlinear change of transmission ratio and stiffness for varying loads and twisting angles \cite{Palli2013}, they have a limited lifetime as repeated cycling will cause creep and wear as fibers tear and stretch, their bandwidth is limited by their intrinsic high reduction ratio and finally, they have limited continuous power output because of the low melting point of the materials currently available for the tendons \cite{Stevens2016}. 

The control of twisted string actuators was thoroughly studied by Palli and colleagues \cite{Palli2013}, who proposed a sliding mode controller using a dynamic model of the twisted strings to achieve high force-tracking accuracy.
A simpler PID feedback controller, closed on a load cell in series with the twisted strings, was used for force tracking in a TSA-driven soft robotic suit for assistance of the elbow joint \cite{Hosseini2020}.

\subsubsection{Shape-memory alloys}
Shape memory alloys are materials with the property of changing shape when subject to an external, typically thermal, stimulus. When the stimulus is removed, the alloy goes back to the original ``memorized shape'' \cite{MohdJani2014}. The most common SMAs, for robustness and thermo-mechanic performance, are Nickel-Titanium (NiTi) alloys, sold in the form of wires that contract when heated and relax when brought back to room temperature. Joule heating is the most common heating method.

SMAs can have a power density of \SI{30}{\watt\per\gram}, over two orders of magnitude higher than the average power density of human skeletal muscles \cite{MohdJani2014} and reach high tensile stresses (up to \SI{200}{\mega\pascal}). Their low-profile, moreover, make SMAs a very appealing solution for actuating soft wearable suits. In \cite{Park2019}, the authors present a soft robotic suit for the elbow joint with a fabric muscle, made from combining 20 SMAs coiled wires in series and parallel, that can exert up to \SI{120}{\newton} of contraction force, has a stroke of 67\% of its resting length and weighs only \SI{24}{\gram}. 

Extensive use of SMA for actuating soft robotic suits is hampered, however, by low efficiency and low bandwidth. The theoretical upper bound for the efficiency of an SMA is the Carnot efficiency between its heating and cooling temperatures, and typical values are below 3\% \cite{MohdJani2014}. Contraction times can be lower than \SI{1}{\second} but relaxation of the same stroke, without an active cooling mechanism, can be as long as \SI{30}{\second} \cite{Park2019}. Active cooling methods such as forced air or immersion in liquids can speed up the cooling process of up to two orders of magnitude \cite{MohdJani2014} but are unpractical in wearable applications. 

Park and colleagues controlled the contraction length of their SMA-based fabric muscle with a PI position controller, closed on a wire encoder in parallel with their SMA-based fabric muscle, and with a low-level current regulator \cite{Park2019}. The exosuit for assisting wrist movements presented in \cite{Jeong2019} and the actuator presented in \cite{Park2019a}, instead, use a feedback temperature controller, closed on a thermocouple attached to the NiTi springs. 

\subsubsection{Pneumatic Interference Actuators}
\label{PIAsection}
When a pressurized bladder with inextensible surface is bent, it tends to buckle in two halves, intersecting at the site of formation of a crease. 
The self-intersection of the pressurize chamber causes a reduction in volume, accompanied by a bending moment proportional to the bending angle, with a spring-like behaviour \cite{Nesler2018}. The torque-bending angle relationship can be modulated by changing the pressure of the air in the chamber. This principle has been used to deliver assistive moments to the human joints, to support, for example, elevation of the shoulder \cite{Simpson2017a, ONeill2020} (Figure~\ref{Upper_limbs}.d), knee extension \cite{Sridar2020} (Figure~\ref{exos_low}.i) or ankle PF \cite{Thalman2019} (Figure~\ref{exos_low}.h). The same principle was used in \cite{Nguyen2019}, to support flexion of the elbow against gravity, where torque was generated by interference of separate, adjacently-placed pneumatic chambers  (Figure~\ref{Upper_limbs}.g).   

PIAs have several advantages that make them an increasingly popular candidate for human movement assistance.  When not-pressurized, they can fold into 2D structures, hardly distinguishable from traditional garments.  Unlike tendons, they apply distributed forces on the human body with a higher component normal to the skin, resulting in better comfort. Moreover, they can be designed to deliver very high torques (up to \SI{324}{\newton\meter} \cite{Veale2020}).
Finally, they are cheap and simple to manufacture: Nesler \etal{} used a Mylar film from a bag of potato chips to line their PIA \cite{Nesler2018}.

While the actuators themselves have a high power-weight ratio, the weight of the compressors or canisters of compressed gas make it unpractical to design an untethered wearable system. Modelling and control of PIA is still an open challenge, due to the complex geometrical and dynamic phenomena that occur in these actuators as they are deformed and pressurized \cite{Nesler2018, Veale2020}.

These non-linear phenomena greatly limit the use of close-loop controllers and affect the accuracy of the open-loop approaches. The latter are currently the most commonly used approach for controlling PIAs, typically using a high-level experimentally-identified mapping between torque, bending angle and pressure and a low-level feedback pressure control, closed on a pressure transducer. The low-level feedback pressure control works by opening and closing of solenoid valves, placed between the source of compressed air and the PIA. Examples of this control approach can be found in \cite{Chung2018,ONeill2017,Sridar2018}.

{\color{black} 
The actuation and sensing modalities used in most of the state of the art soft robotic suits exploit existing mature technologies, often borrowed from other fields. The most promising effects on human motor performance were achieved with the electric motor tendon units. The main reason lies in the maturity of electromagnetic  motors, with a consequent large body of knowledge on how to control them and in the availability of cheap and easy-to-use electronics to do so. For researchers and developers interested in a quick transition from an idea to a working prototype, this is the fastest and easiest route. The recent interest in electric vehicles, moreover, has promoted the development of cheaper and more performing electric motors from the drone and automotive industry. These advantages are likely to improve accessibility and efficiency of soft robotic suits in the coming years. Outrunner electric motors from the drone industry, for example, having high-torque density and efficiency, are interesting candidates for wearable applications \cite{Lee2019}.

As the interest in soft robots grows, however, new, \textit{ad hoc} actuation \cite{Zhang2019a} and sensing modalities \cite{Lu2014} are emerging, suggesting that, in the long-term, soft robotic suits will look rather different.

Exomuscles of the future may supplant their motors and encoders, or compressors and gauges, with ``smart textiles''\cite{Sanchez2020}; revolutionary materials that incorporate both actuation and sensing into a completely soft, flexible garment. A growing and promising interdisciplinary collaboration between roboticists and material-scientists has led to the recent development of many novel actuation techniques: it might soon be more common to find a suit driven by wireless liquid-gas phase-changes \cite{Boyvat2018}, or electrochemical reactions than a motor-tendon pair. 
}
\subsection{Physical human-suit interface}\label{pHRI}

 {\color{black} The most defining characteristic of soft robotic suits is the way they transfer forces to the human body. While their actuators and sensing strategies still rely on mature technologies such as electric motors, McKibben muscles, IMUs and pressure sensors, the use of fabric and elastomers to create load-paths from the actuators to the human skeleton is a young and unexplored field. More and more research labs are being populated with sewing machines, as fabric design and textile engineering become paramount for the effectiveness of exomuscles. }
 
 It is key to face the challenge of finding an efficient and comfortable mechanical means to transmit forces from a robot to the human body.  { \color{black} This is all the more important for devices intended for people with sensory impairment, where, without the user's feedback, the robot could apply high pressures for long periods of time, eventually leading to pressure ulcers or local blood flow obstruction \cite{Kermavnar2018b}.} 
 \begin{figure}[htbp]
    \centering
    \includegraphics[width = .3\textwidth]{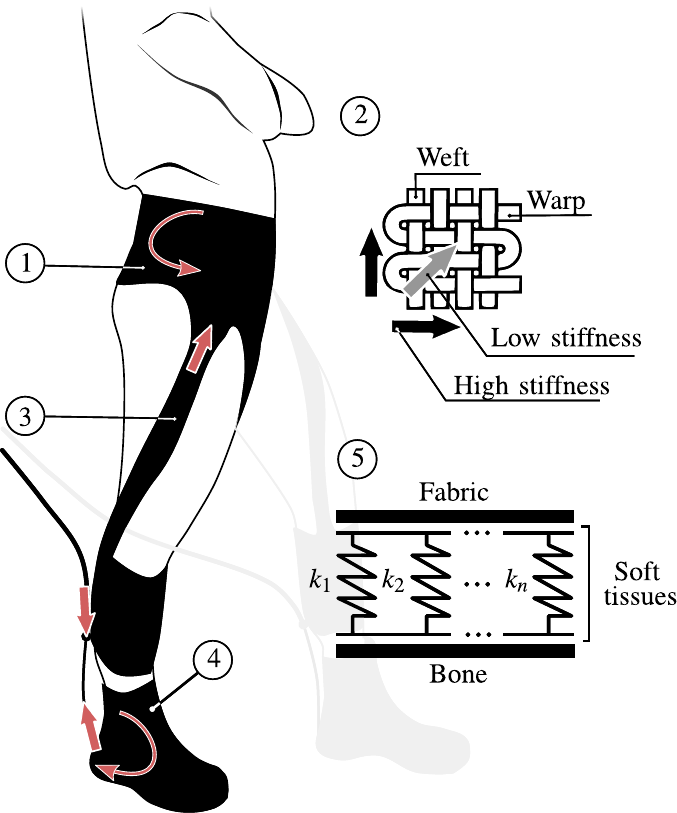}
    \caption{\color{black} Practical guidelines for designing the pHRI of a soft robotic suit, and their embodiment in the Harvard exosuit \cite{Asbeck2015}. \textcircled{1} Anchor the suit to locations on the body that have high stiffness, such as the iliac crest. \textcircled{2} Choose suit materials with as high a tensile stiffness as possible; most fabrics are stiffer along the lines of their weft and warp. \textcircled{3} Follow the most direct paths for transferring force across the body. \textcircled{4} Minimize shear forces between the suit and human; using shoes is a clear example, as forces can be transferred normally through the sole. \textcircled{5} Maximize the suit–skin contact area. Covering a larger area with the fabric achieves higher transmission stiffness because, as the interface surface increases, the underlying soft tissues work as parallel springs. Red arrows indicate paths of the actuating forces and resulting moments around human joints on the sagittal plane. }
    \label{fig_pHRI}
\end{figure}

De Rossi \textit{et al.} proposed an apparatus to monitor the distribution of pressure at the human-robot interface of a lower-limb exoskeleton during gait training \cite{DeRossi2011}. Using a similar approach, Levesque \textit{et al.} identified areas exhibiting peaks of pressure on the attachment points of an active orthosis for gait assistance \cite{Levesque2017}; their findings suggest that the distribution of pressure is affected more by the stiffness of the padding material than by its thickness.

This data-driven approach to guide design choices is a promising path for soft robotic suits: Quinlivan \etal{} used the same paradigm to optimise the topology and material composition of the attachment points of the Harvard exosuit \cite{Quinlivan2015}. This work highlighted the importance of the geometry of the interface  as well as advising the use of fabric materials that better conform to the human body. Yandell \etal{} analysed the maximum tolerated forces that can be applied to the shoulders, thighs and shank, finding that humans better tolerate forces that are applied faster, that there is a strong and significant habituation mechanism and that a tendon-driving mechanism can comfortably transmit up to \SI{950}{\newton} on healthy people \cite{Yandell2020}. 

A recent meta-analysis on cuff pressure algometry \cite{Kermavnar2018} suggests that exposure to ``circumferential compression leads to discomfort at approximately 16-\SI{34}{\kilo\pascal}, becomes painful at approximately 20-\SI{27}{\kilo\pascal}, and can become unbearable even below \SI{40}{\kilo\pascal}''. These limits are much lower, however, in people with chronic pain, indicating the need for much more stringent requirements for soft robotic suits in the healthcare domain: the threshold for pain is between 10 and \SI{18}{\kilo\pascal}, and pressures can be unbearable even below \SI{25}{\kilo\pascal} \cite{Kermavnar2018}.

These studies focused on pressure, but there is evidence that forces tangential to the skin are no less important in affecting comfort and blood flow \cite{MingZhang1993}. Most soft robotic suits in the literature, especially when driven by tendons, strongly rely on friction between the skin and the cuff to transfer power to the human body. However, methods to measure shear forces and solutions to reduce them are still very scarce. This problem has been addressed in studying forces at the interface of a prosthetic socket \cite{Zhang1998} and only recently was a method proposed, to measure both shear and normal forces simultaneously between a textile cuff and the skin \cite{Georgarakis2018}.

To complicate things further, the requirement of comfort is often at odds with that of efficient transfer of power between the suit and the human body: a stiff transmission is more mechanically efficient but often not comfortable. Yandell and colleagues indeed found that the dynamics for the interface between a soft suit and the actuators have significant implications for the timing, magnitude and efficiency of assistance \cite{Yandell2017}. Soft tissues, fabric and flexible transmissions are huge power sinks, absorbing up to 55\% of the actuator power. Compliance of the transmission and interface, moreover, adds a time delay in the transfer of force from the actuator to the human skeleton. If not accounted for, it negatively affects timing and, consequently, effectiveness of assistance \cite{Yandell2017}. 

When developing the Harvard exosuit, the team from the Biodesign lab formulated a very helpful set of practical guidelines to consider when designing the fabric components of a wearable robot. These general principles have the goal of maximizing power transfer (1-3) and comfort (4-5) of the device \cite{Asbeck2015} {\color{black} (Figure~\ref{fig_pHRI})}. 

Technological solutions to tackle these challenges are, however, still limited, probably slowed by the inertia in accepting that textile research is a fundamental discipline of wearable robotics.
SRI spin-off Superflex, shown in Figure~\ref{exos_low}.f, designed a cuff for anchoring their actuator to the calf that exploits the topology of webbing braids to distribute forces evenly on the lower leg. The FlexGrip technology 
functions in a manner similar to a Chinese finger trap: the cylindrical cuff is made of a helically-wound webbing braid that, upon the application of a force parallel to its main axis, contracts radially. The contraction is caused by a reduction in the angle between the warp and weft threads at their crossing points \cite{Witherspoon}. The result is a more uniform pressure distribution on the soft tissues under the cuff that flattens peaks of pressure and increases transmission stiffness.

This idea was taken a step further by Choi and colleagues \cite{Choi2019}, who proposed an active cuff for a soft robotic suit for assisting the wrist.  Using a tendon-pulley mechanism with a small electric motor, the cuff compresses the underlying tissues when the robotic device delivers assistance, to ensure efficient transfer of forces, and releases the pressure when the device is idle.

The use of soft materials has bypassed two of the most prominent problems of the pHRI between a wearable robot and its user. Firstly, soft robotic suits do not suffer from  kinematic incompatibility, known to cause misalignment, parasitic interaction torques and restricted ROM. Moreover, the mass of their textile anchor points or inflatable actuators is often negligible compared to the one of the human limbs, allowing for the device to become dynamically transparent by simply slacking a cable or venting an inflatable chamber. If exomuscles are to become daily assistive devices, however, there is a dire need for efforts towards improving their comfort. 


To bridge this gap, we need to better understand the factors that affect comfort of a device that exerts forces on the human body. Existing literature from other fields suggest that this is a complex interplay between topology and material of the interface components, microclimate of the skin \cite{Zhong2006}, superficial capillary blood flow and magnitude of external forces --- both normal and tangential to the surface of the body \cite{L1979}. Data-driven approaches, combined with subjective feedback, will hopefully inform the choice of materials, the design and technological solutions to allow for a robotic suit to be comfortably worn for a full day without causing discomfort, blisters or pressure ulcers.

Technological innovations also need to contribute towards this goal. A promising starting point could be the adaptation of technological solutions from different fields that face the same problem: Sengeh \etal{}, for example, designed a prosthetic socket for transtibial amputees with variable 3D stiffness, designed to mirror the stiffness of the residual limb and avoid peaks of pressure \cite{Sengeh2013}. Active cuffs \cite{Choi2019} or designs that exploit a smart topology of the fabric fibers to uniformly distribute forces on the underlying soft tissues \cite{Witherspoon} are also auspicious paths. 

\subsection{Sensors and strategies for intention-detection}\label{cHRI} 
In \cite{Pons2008}, J. L. Pons nicely points out that the human-robot interface (HRI)  between a wearable robot and its user consists of processes of two different natures: (1) the physical interface (pHRI) that we discussed in Section~\ref{pHRI} and (2) the exchange of cognitive information between the two agents (cHRI).

For correct functioning of a wearable robot, the user's intentions (cHRI) needs to be understood and assisted by the device in an accurate and timely manner. This is all the more important for soft robotic suits, designed to work in concert with their wearer rather than impose a movement. 

Sensors and methods to infer the intention of the user can be broadly classified in those that make use of a mechanical manifestation of movement and those that make use of neural signals. The former employ transducers of kinematic and kinetic aspects of movement; typical examples are load cells, pressure sensors, encoders, inertial measurements units and, more recently, flexible stretch and strain sensors. Controllers that make use of such sensors to decode human intention are also known as ``mechanically intrinsic controllers''. Neural controllers, instead, rely on signals from the human central or peripheral nervous system, acquired, for example, through surface ElectroMyoGraphy (sEMG).

Theoretically, capturing information from neural signals would be the ideal case: the efferent control command generated by the brain to activate muscles precedes movement and adapts to changing environmental dynamics, allowing the robot to apply forces similar, in timing and magnitude, to those of the muscles it is working in parallel with.  The soft robotic suit for elbow assistance presented in \cite{Hosseini2020}, recently used sEMG signals from the biceps and the triceps brachii, to guide a state-machine that supports lifting tasks. Lotti and colleagues \cite{Lotti2020,Missiroli2020}, instead, drove an exosuit for the elbow joint using an accurate model of the geometry and activation dynamics of the biceps brachii, triceps brachii and brachioradialis, to map sEMG to a desired assistive torque \cite{Sartori2012}.

Neural controllers, although in theory very promising, are typically more complex than mechanically intrinsic ones, due to the noisy nature of neural signals, variable impedance properties of the skin, sensitivity to muscular fatigue and the need for frequent calibration \cite{Merletti2004}.
For this reason, most of the soft robotic suits in the literature use mechanically intrinsic control approaches. Although these strategies are intrinsically prone to a time delay between the initiation and detection of human movement, they rely on robust and repeatable measurements.

Mechanically intrinsic controllers for soft robotic suits that assist the lower limbs are almost all based on gait-event detection. This is typically done with a foot-switch or pressure sensors beneath the feet (e.g., \cite{Park2014,Thalman2019}) or IMUs \cite{Kim2019, Awad2017, Haufe2019}, placed on the thighs, shanks and/or feet. Signals from these sensors are used to detect gait events such as heel-strike and toe-off, in order to segment steps and phases within each step and trigger a desired assistive profile. 


The downside of mechanically-intrinsic controllers is limited adaptability to varying conditions. Existing controllers, for example, do not adapt magnitude of assistance to the different dynamic demands of walking at varying speeds, slopes and carried loads; none of the available controllers developed so far can discern and adapt to tasks such as jumping, stair climbing or walking downhill. This lack of adaptability limits the benefit of the device, in the best case and can restrict movements and induce falls in the worst case. 

What has been robustly achieved is adaptation of the timing of assistance to varying walking cadence. In \cite{Asbeck2015}, the authors attain this by adjusting the gait period of the assistive profile to the average period of the last five steps. A more responsive approach is proposed in \cite{Asbeck2015a}, using the time delay between two different known events within the same gait cycle.

Finally, Kim \etal{} used the vertical acceleration of user's Center of Mass (CoM) at the moment of maximum hip extension to discern walking from running with a 100\% accuracy, and trigger the appropriate assistance profile for each mode of locomotion \cite{Kim2019}. Both the acceleration of the CoM and the thigh angles were obtained by IMUs, placed on the abdomen and thighs. 

Fewer works are available on mechanically-intrinsic intention-detection strategies for soft robotic suits that assist the upper limbs. This is probably due to the challenge introduced by the higher variability of volitional movements of the arms, fundamentally different from the rhythmic nature of walking. A commonly used approach is a switch or manual input that triggers a desired force or pressure from the actuator, independently of the movements of the wearer. This is the case, for example, for the Harvard shoulder exosuit \cite{ONeill2020, ONeill2017}, the work from Abe \etal{} \cite{Abe2019} and the device for supporting elbow movements described in \cite{Nguyen2019}. This simple approach is limited but practical and could be sufficient for most tasks in daily life that involve movements against gravity, especially when the actuator is a PAM or PIA with a ``forgiving'' intrinsic spring-like behaviour.

A more complex strategy was proposed in \cite{Chiaradia2018}, for a tendon-driven device, to compensate for gravitational forces acting on the elbow joint. The device used a commercially available capacitive stretch sensor (StretchFABRIC, Stretchsense, New Zeland), aligned with the human joint, to monitor the elbow angle. 


{\color{black} 
Means to understand the intention of the user and the environmental conditions in which he/she is moving are still far from optimal. 
The uptake of soft robotic suits  will very likely depend on the design of intuitive control approaches that do not impose additional cognitive load on their users. 
To shape a future where soft robotic suits will help us to walk, run, jump, dance and restore independence in people with neuromuscular or neurological impairments, we need to work towards improved versatility and personalization of high-level controllers for intention-detection. 

The available intention-detection algorithms for lower limbs  exploit the rhythmic nature of locomotion to detect events and deliver assistance with the right timing. Their control laws, however, make stringent assumptions about the environmental conditions and the tasks that the wearer performs: for example, current assistive profiles neither change in magnitude with the speed of walking nor do they automatically adapt to accommodate for inclined or declined surfaces or diverse tasks such as sit-to-stand transitions, stair-climbing or jumping. We need more efforts to improve the versatility of soft robotic suits for the lower limbs, either through mechanically-intrinsic controllers (e.g. the IMU-based algorithm to discern walking from running proposed in \cite{Kim2019}), robust neural controllers that leverage on the adaptability of the human nervous system to drive robotic assistance (e.g. \cite{Lotti2020}), or even emerging vision-based approached \cite{Kim2019a}. 

The problem of adapting to changing task and dynamic conditions is even more challenging for the upper limbs. In most of the studies presented in this review, control of the user over the device was limited to a manual input, to turn the assistance on/off. This approach is extremely robust but places an additional cognitive burden on its user. 
\begin{table}[htbp]%
    \small
	\caption{Common benchmarking metrics for soft robotic suits}
    \centering
    \begin{tabular}{l| c}
    \hline
    \hline
    Metric   & Unit  \\ \hline
    \rowcolor{mygray}
    \textbf{Cardiovascular function} &   \\
    \hspace{3mm} Net metabolic rate  & \SI{}{\watt\per\kilo\per\gram} \\
    \hspace{3mm} Cost of transport  & - \\
    \hspace{3mm} Heart rate  &beats min$^{-1}$  \\
    \rowcolor{mygray}
    \textbf{Biomechanics} &   \\
    \textit{Surface EMG} &  \\
    \hspace{3mm} Amplitude & \SI{}{\%MVC} or \SI{}{\milli\volt}  \\
    \hspace{3mm} Median frequency  & \SI{}{\hertz}   \\
    \textit{Kinetics}  &    \\ 
    \hspace{3mm} Joint torques  & \SI{}{\newton\meter}  \\
    \hspace{3mm} Joint powers  & \SI{}{\watt} \\
    \hspace{3mm} Robotic forces/torques  & \SI{}{\newton} or \SI{}{\newton\meter}   \\
    \textit{Kinematics}  &    \\ 
    \hspace{3mm} Joint angles  & \SI{}{\degree}  \\
    \hspace{3mm} Range of motion  & \SI{}{\degree}  \\
    \textit{Spatio-temporal parameters}  &    \\ 
    \hspace{3mm} Smoothness of movement & -  \\
    \hspace{3mm} Speed of walking & \SI{}{\meter\per\second}  \\
    \hspace{3mm} Duration of gait phases & \SI{}{\%}  \\
    \hspace{3mm} Symmetry of walking & -   \\
    \rowcolor{mygray}
    \textbf{Psychophysiology}  &    \\ 
    \hspace{3mm} Borg scale rating  & 15 point scale  \\
    \hspace{3mm} Qualitative questionnaires & VAS$^*$ or Likert scale \\
    \hline
    \hline
    \end{tabular}
  \\ \raggedright
    \label{table_benchmarking}
    \footnotesize{$^*$ Visual analog scale.}
\end{table}
Another pressing issue when designing a controller for cHRI is that of personalization. In spite of some very well-known stereotypical patterns, human movement has some distinctive idiosyncratic features --- we can recognize someone simply by the way they walk \cite{Cutting1977}! This has motivated the design of online optimization techniques that personalizes the characteristics of assistance of a wearable robot (e.g., \cite{Zhang2017c,Ding2018, Haufe2020d}), in order to minimize the wearer's metabolic rate. This approach lead to some of the highest achieved metabolic benefits during locomotion, e.g. \cite{Lee2018}. 

}


\section{Effects on human movements}
In a distinguished review on the state of the art of lower limb exoskeletons, in 2008, Dollar and Herr highlighted the lack of published quantitative results on the effectiveness of exoskeletons \cite{Dollar2008}. This is not the case for soft robotic suits now. Their simplicity and and affordability has allowed for a faster and more practical human-in-the-loop design process \cite{Walsh2018}. This human-centered validation process has resulted in a comprehensive body of literature on the effect of robotic suits on human movements, using a relatively small set of common outcome measures. The most commonly used ones are shown in Table~\ref{table_benchmarking}. 
 \begin{figure}[htbp]
    \centering
    \includegraphics[width = .5\textwidth]{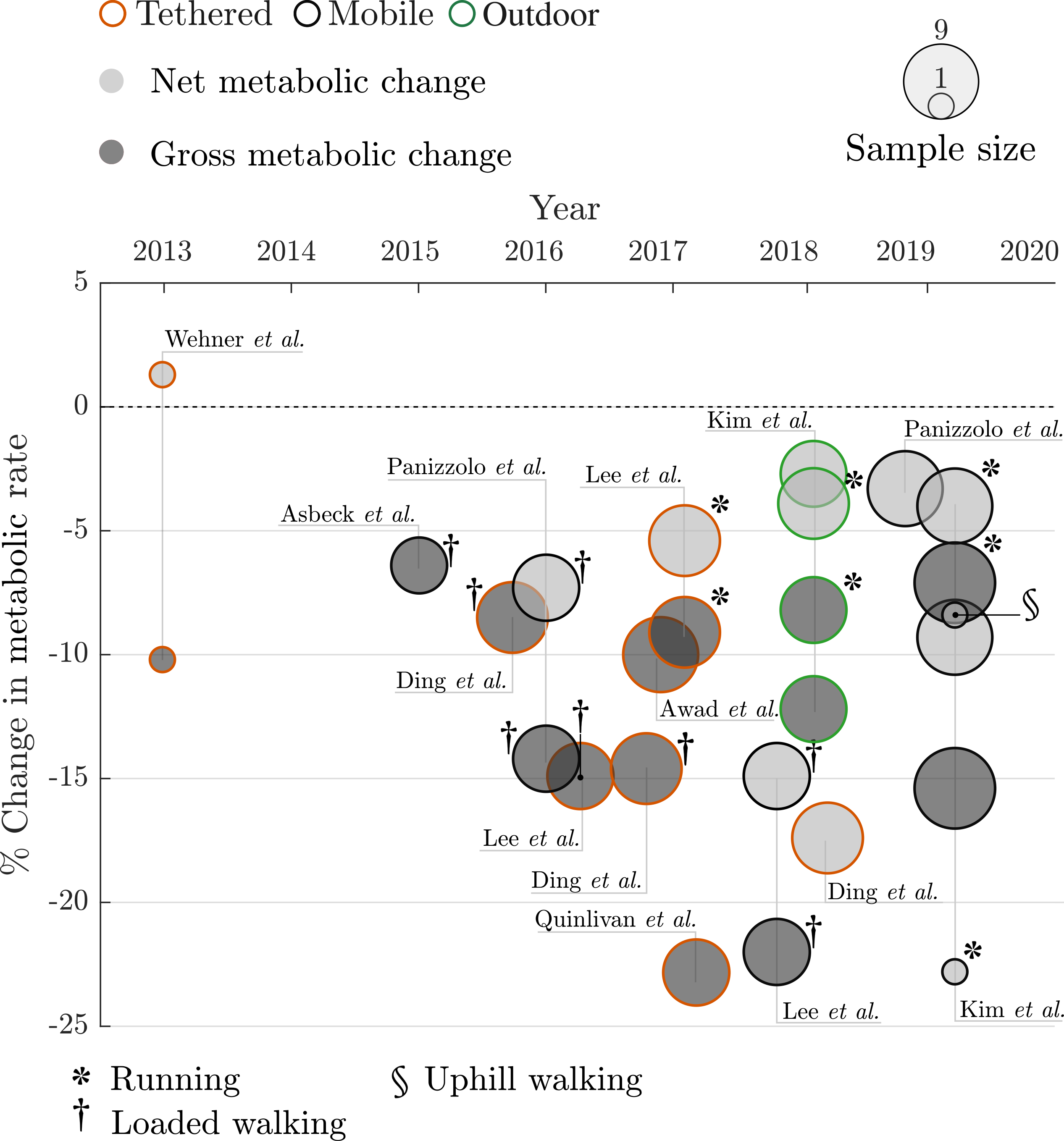}
    \caption{Studies reporting the change in Metabolic Rate (MR) related to walking or running, resulting from receiving assistance from a soft robotic suit. This representation is not meant to compare study outcomes, as device characteristics and experimental conditions vary across studies.}
    \label{Metabolics}
\end{figure}
\subsection{Soft robotic suits for lower limbs}
When it comes to locomotion, the performance of a soft robotic suit is most commonly measured by the relative change it induces in the  metabolic rate (MR) of its wearer, defined as the metabolic rate during locomotion minus the rate for quiet standing. MR is a comprehensive measure of the device performance, reflecting both its technical prowess and its complex interaction with the human agent.

Figure~\ref{Metabolics} shows the percentage change in MR resulting from receiving assistance from a soft robotic suit for the lower limbs during walking or running. We show both studies that report the gross and the net metabolic change; these are defined adopting the convention used in \cite{Lee2018}: 
\begin{align}
    \text{Net metabolic change} &=  \frac{ MR_{Powered} - MR_{No\_suit}}{MR_{No\_suit} - MR_{Standing}} \\
      \text{Gross metabolic change} &=  \frac{MR_{Powered} - MR_{Unpowered} }{MR_{Unpowered}- MR_{Standing}}.
\end{align}
The former reports the relative effect of receiving assistance from the soft robotic suit, compared to walking without wearing the device. The gross metabolic benefit, instead, uses a condition where the device is worn but turned off as a reference, resulting in a less realistic but experimentally more practical metric, often used to compare the effect of control or hardware changes (e.g. \cite{Quinlivan2017}).

The other important dichotomy worth highlighting here is that between mobile devices and robotic suits tethered to an off-board actuation platform. The change in MR resulting from wearing a tethered device does not reflect the metabolic penalty of the mass of the off-board components. Tethered devices are very common in research laboratories because placing the actuator on an off-board platform allows for cheaper, faster and more versatile prototyping \cite{Witte2020}. Using a tethered device is also more convenient for physical therapy in controlled environments \cite{Bae2015}.

{\color{black} 
The first tethered prototype of the Harvard exosuit (Figure~\ref{exos_low}.a), dating back to 2013 and using a contractile PAM for ankle PF, achieved a gross metabolic change of -10.2\% and a net metabolic change of 1.3\%, in a single-case study \cite{Wehner2013}.
Metabolic benefits for human walking have since been substantially been improved, with a peak gross change of -22.8\% \cite{Quinlivan2017}.

Most of work in the literature focused on more power-demanding tasks such as loaded walking , with both tethered \cite{Ding2016, Ding2017} and mobile \cite{Asbeck2015,Panizzolo2016,Lee2018} devices. The highest achievements for loaded walking are a gross change of -22\% \cite{Lee2018} and  net change  of -14.9\% \cite{Lee2018}. For comparison, a net metabolic change of 20\% is approximately equivalent to the effect of carrying a \SI{25}{\kilo\gram} backpack. 

Soft robotic suits can also support running: a device for hip extension was shown to improve the economy of human running (\SI{2.5}{\meter/\second}), with a net metabolic change of -4\% and a gross change of -9.1\% \cite{Lee2017b,Kim2019}. The highest achieved reduction with a wearable robot for human running belongs to Witte \etal{}, with a \SI{-14.6}{\%} net metabolic change at a speed of \SI{2.7}{\meter/\second}, using a rigid exoskeleton for ankle PF, controlled by an off-board actuation platform \cite{Witte2020a}. For comparison, the Nike Vaporfly shoes, used to break the world record in \SI{100}{\kilo\meter}, marathon, half-marathon and \SI{15}{\kilo\meter} in 2018, were shown to reduce the cost of running at \SI{3.9}{\meter/\second} by \SI{4.16}{\%} compared to established racing shoes \cite{Hoogkamer2018}. 
}


Soft robotic suits have been shown to provide benefits also for people with mobility impairments. A unilateral tethered version of the Harvard exosuit for ankle PF/DF (Figure~\ref{exos_low}.c) was shown to significantly improve ankle swing DF and forward propulsion of the hemiparetic leg, in nine stroke survivors \cite{Awad2017}. This resulted in a more symmetric and efficient gait pattern and a corresponding gross metabolic change of -10\%. A more recent multi-site clinical trial proved that a similar commercialized device is a safe and feasible means of supporting post-stroke gait rehabilitation \cite{Awad2020a}.

Similarly, the Myosuit was shown to improve walking speed of a person with incomplete spinal cord injury, by a clinically-meaningful \SI{0.16}{\meter/\second} \cite{Haufe2019} and reduce the cost of transport by 9\%, compared to a no-suit condition, on an outdoor sloped mountain path \cite{Haufe2020a}. In a feasibility study with eight participants, moreover, Haufe \etal{} showed that the Myosuit is a safe and practical tool for physical training with people suffering from a variety of conditions, including stroke hemiparesis, iSCI and muscular dystrophy \cite{Haufe2020b}.  

Finally, the Exoband achieved a reduction in net metabolic cost of walking of 3\% in an population of older adults, simply using elastic bands that support limb advancement \cite{Panizzolo2019a}. This study, together with \cite{Haufe2020c}, once again emphasize the importance of carefully considering the passive dynamics of a wearable device in its design process.
 \begin{figure}[htbp]
    \centering
    \includegraphics[width = .45\textwidth]{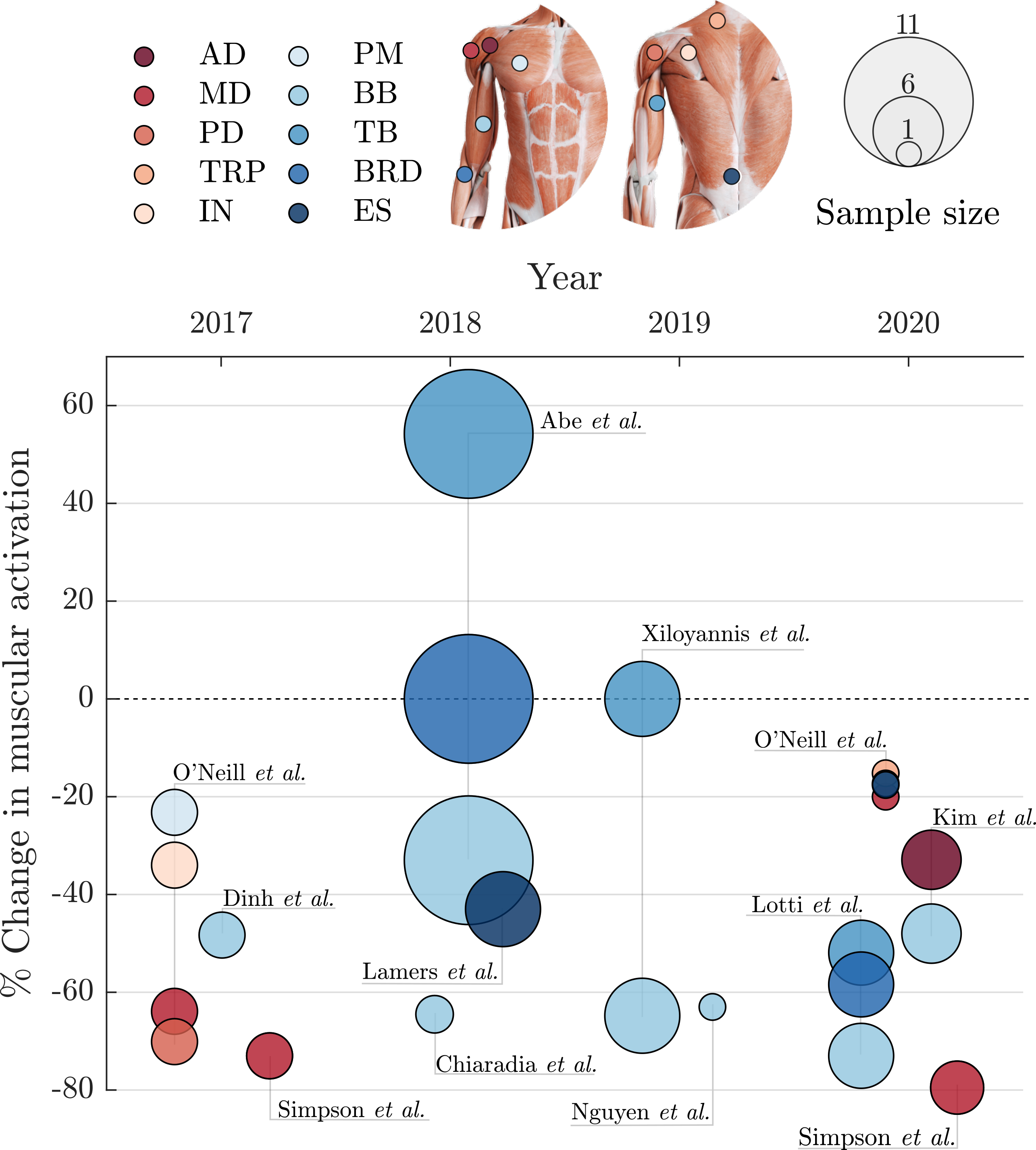}
    \caption{Studies reporting the change in muscular effort, resulting from wearing a soft robotic suit that assists movement of the upper limbs or the trunk. Size encodes sample size and color encodes the muscle whose activity was reported.  This representation is not meant to allow for comparison of devices, as experimental conditions vary across studies. Acronyms: AD anterior deltoid; MD medial deltoid; PD posterior deltoid; TRP trapezius; IN infraspinatus; PM pectoralis major; BB bicpes brachii; TB triceps brachii; BRD brachioradialis; ES erector spinae.  }
    \label{f_EMG}
\end{figure}

\subsection{Soft robotic suits for the upper limbs and trunk}
Studies reporting the effect of soft robotic suits for the upper limbs on human movements are fewer and rely on a less homogeneous set of outcome measures. A commonly-used metric is the relative change in muscular activity resulting from wearing the device, extracted from the amplitude of surface EMG activity of selected muscles, affected by the support from the robot.

Figure~\ref{f_EMG} shows the relative reduction reported in studies that investigate the effect of soft robotic suits for the upper limbs, with emphasis on the muscles monitored and sample size. We adopt the terminology introduced for the metabolic results and refer to a net reduction when the control condition is a ``no suit'' one, and gross reduction when results are reported as a percentage of a ``suit on but unpowered'' case. Unlike for studies on locomotion, most of the devices for the arms target people with mobility impairments. 

O'Neill and colleagues assessed the effect of a soft PIA-actuated exosuit for supporting shoulder abduction and horizontal F/E, reporting a gross reduction in the activity of the medial deltoid (MD), posterior deltoid (PD), infraspinatus (IN) and pectoralis major (PM), in a cohort of three healthy participants \cite{ONeill2017}. 
A re-iteration of the same device (shown in Figure~\ref{upper_limbs}.d) was used as a means to aid both severe stroke patients and therapists during physical rehabilitation \cite{ONeill2020}. 


Similarly, Simpson \etal{} proposed a PIA-based device and reported a net change of -79.5\% in the anterior deltoid (AD) and MD, on four healthy participants and an improved workspace area with chronic stroke patients \cite{Simpson2020}. 


One of the highest achieved reductions in the muscles around the elbow joint belongs to Lotti \etal{}, using an EMG-based controller \cite{Lotti2020}. Here the device achieved changes of up to -73\%, -77\% and -59.7\% for the BB, brachioradialis (BRD) and triceps brachii (TB), respectively. 

Lightweight unobtrusive devices to support the upper limbs and trunk are an appealing solutions for assisting repetitive lifting movements in industrial scenarios.
Kim and colleagues achieved a -48\% and -32\% gross reduction in the activity of the BB and AD, respectively, with a soft robotic suit using an electric motor to drive a tendon routed on a bi-articular path across the shoulder and elbow joint \cite{Kim2020}. 
Nguyen \etal{} showed a higher reduction in the activity of the BB, during an isometric task (-63\% gross change) using an array of expansive PIAs, mounted on the elbow joint to support flexion movements (Figure~\ref{Upper_limbs}.g) \cite{Nguyen2019}.

Finally it is worth mentioning results achieved with a very simple passive suits \cite{Lamers2018} (shown in Figure~\ref{Upper_limbs}.e), using elastic bands to unload the erector spinae muscles. The authors found a peak net change on the activity of the muscles group of -43\%, during a leaning task.

\vspace{3mm}
{\color{black} 
In a provoking article from 2007, Ferris \etal{} encouraged robotic engineers to work more closely with physiologists and movement scientists to fine-tune the design process and  use wearable robots as tools to investigate fundamental scientific questions on human neuromotor control \cite{Ferris2007}. One of the key messages of Ferris' script, i.e. to put more emphasis on the reduction of metabolic cost as a key performance metric for wearable robots, has been heard clearly by the community \cite{Sawicki2020}.
The other encouragement was that of understanding more about the neural mechanisms involved in the use of such devices.

How long does it take for a healthy user to learn to fully benefit from the use of a soft robotic suit? What neural processes are involved? How do neuromotor impairments affect these mechanisms? Recent work has started to scratch the surface of these questions \cite{Panizzolo2019}. 
The mechanical transparency, simplicity and versatility of soft robotic suits make them ideal tools to address these questions. The answers will inform the invention of better assistive devices, help to better understand motor learning and recovery processes, conceive scientific protocols that account for adaptation times and contribute to manage end-users' expectations. 
}


\section{General conclusions}
{\color{black} 
Through their simplicity and portability, soft robotic suits have brought wearable robots a step closer to our daily lives. Already a couple of research groups have stepped outside of research laboratories, clinics and hospitals to test the effect of robotic assistance in daily environments, e.g. on rough terrains \cite{Kim2018a} (Figure~\ref{Metabolics}, in green) and sloped mountain paths \cite{Haufe2020a}.  

Future studies and wider spread of this technology will give us insights on the effect of soft robotic suits in  outdoors and domestic settings. Moving out of the confinement of laboratories will allow us to answer questions on the functional impact of this technology on people's lives.

As we do so, we need to face and solve some critical technological challenges:
\begin{itemize}
    \item {\color{black} First and foremost, we still do not know how to attach a robot and transfer forces to the human body. This is an under-addressed topic but one in need of urgent systematic investigation and original ideas.}
    \item The actuation landscape is still largely dominated by an almost two-century-old technology, the electric motor. Efforts to solve the control challenges of PAMs and PIAs and new means of delivering pneumatic energy (e.g. chemical or electrochemical reactions) will have a lasting impact on the field. 
    \item As wearable robots populate our daily environments, we will need them to be increasingly versatile, designing high-level controllers that adapt and seamlessly transition between diverse tasks and dynamics. For the lower limbs, these include walking, stairs ascend/descend, jumping, sitting and running. For the upper limbs, this includes adaptation to changing loads, and speeds of movement.
\end{itemize}

Once we tackle these obstacles, soft robotic suits will have a concrete impact on society: they will improve mobility in healthy young adults, reduce the likelihood of injury in industry, help elderly to keep up with their younger peers and support people with motor disabilities in living a more active and fulfilling life. 
}

\ifCLASSOPTIONcaptionsoff
  \newpage
\fi

\end{document}